\documentclass[letterpaper]{article} % DO NOT CHANGE THIS
\usepackage[submission]{aaai2026}  % DO NOT CHANGE THIS
\usepackage{times}  % DO NOT CHANGE THIS
\usepackage{helvet}  % DO NOT CHANGE THIS
\usepackage{courier}  % DO NOT CHANGE THIS
\usepackage[hyphens]{url}  % DO NOT CHANGE THIS
\usepackage{graphicx} % DO NOT CHANGE THIS
\graphicspath{{figs/}}
\usepackage{natbib}  % DO NOT CHANGE THIS AND DO NOT ADD ANY OPTIONS TO IT
\usepackage{caption} % DO NOT CHANGE THIS AND DO NOT ADD ANY OPTIONS TO IT

\usepackage{array}
\usepackage{booktabs}
\usepackage{amsmath}

\usepackage{xcolor}

\usepackage{listings}
\nocopyright %-- Your paper will not be published if you use this command
\usepackage{multirow}
\usepackage{arydshln} 
\usepackage{amsmath}
\usepackage[show]{notes}
\newcommand{\ignore}[1]{}
\newcommand{\com}[1]{}

\newif\ifshowcomments
\showcommentsfalse

\ifshowcomments
    \newcommand{\oren}[1]{\inote[oren]{\textcolor{blue}{\bf #1}}}
    \newcommand{\lior}[1]{\inote[Lior]{\textcolor{cyan}{\bf #1}}}
    \newcommand{\obj}[1]{{\textcolor{purple}{\bf #1}}}
    \newcommand{\gap}[1]{{\textcolor{red}{\bf #1}}}
\else
    \newcommand{\oren}[1]{}
    \newcommand{\lior}[1]{}
    \newcommand{\obj}[1]{}
    \newcommand{\gap}[1]{}
\fi

\title{Agentic Detection of Online Conspiracies}
\author{
    Lior Biton~~~~~ Oren Tsur
}
\affiliations{
    Department of Computer and Information Science\\
    Ben-Gurion University of the Negev, Israel\\
    \texttt{bitol@post.bgu.ac.il ~~~~~orentsur@bgu.ac.il}
}

\begin{document}

\maketitle

\begin{abstract}
Conspiratorial discourse on social media is not always expressed through explicit claims or stable lexical markers. The same surface content may express endorsement, legitimate concerns, criticism, satire, or mockery. The main challenge is therefore not only recognizing conspiracy-related claims, but inferring the speaker's intent -- the utterance's illocutionary force. We argue that this can be achieved through the use of relevant social contexts and propose an agentic framework, equipped with a set of tools supporting social queries. 

We demonstrate the benefits of our approach on a unique dataset of Hebrew tweets, covering 80\%--90\% of the public Hebrew tweets published over a four-year span (late 2018-- early 2023), encompassing several election cycles as well as the COVID pandemic years and related vaccination campaigns. This extensive coverage can be used in recovering different social contexts. 
Evaluating our framework on a manually-annotated adversarial dataset, we find that context-aware workflows consistently outperform text-only classification and that the agentic framework performs better than other frameworks and settings, with significant gains over a non-agentic model exposed to the same contexts available to the agent. 
We further provide an analysis of the results, the errors and efficiency (token economy) tradeoffs. 

These findings support viewing the task of conspiracy detection as a socially embedded interpretation task, in which effective classification depends not only on access to contexts, but also on adaptive reasoning in which the agent uses tools on a per-case basis, asking only for evidence relevant to its current reasoning step. 
\end{abstract}

\section{Introduction}
\label{sec:intro}

Social media has become a central arena for the circulation of misinformation, rumors, and conspiracy theories \cite{grinberg2019fake,muhammed2022disaster}. %These narratives do not spread only through isolated false claims or automated accounts. T
Commonly referred to as \emph{alternative narratives} \cite{starbird2017examining}, they are spread and consumed through an ecosystem that combines influencers with various motivations,\footnote{Motivations range from pure monetary gain, to the perception of self, sense of belonging, or goals of nation-state actors.} committed communities, and the participation of ordinary users \cite{starbird2019disinformation}. 
These narratives tend to gain traction beyond fringe communities, especially in times of declining trust in institutions: increased political polarization, social unrest, economic or public-health crises, and wars -- times in which these `bespoke realities' are ever more dangerous \cite{douglas2023conspiracy,allen2024quantifying,diresta2024invisible,hilberts2025impact}. Support-for and participation-in political violence were found to be correlated with beliefs in conspiracy theories, especially those held by small homogeneous groups \cite{enders2024relationship}.
The analysis of the patterns of conspiratorial discourse is of great interest to academic scholars, policy-makers, and the online platforms serving as the main vehicle for the phenomena. 
However, the detection of conspiratorial discourse is theoretically and algorithmically
challenging \cite{mahl2023conspiracy}. 
Conspiratorial discourse is not well defined. The same surface-level text may serve different communicative acts, ranging from false narratives to legitimate concerns\footnote{E.g., ``\emph{Is Tylenol safe during pregnancy?}''}, and mockery.\footnote{E.g., ``\emph{The COVID vaccine is designed to be activated by a 5G cellular signal, triggering a tail growth}'' and see Fact Check: COVID-19 vaccines are not a ploy to connect people to 5G; Accessed 7/30/26.}
The challenge is more pronounced in short and informal communication styles prevalent in some online social platforms, and among members of cohesive communities using in-group jargon. 
Moreover, challenging a dogma with alternative narratives can prove true,\footnote{Consider the legacy of Copernicus, Galileo Galilei, or Woodward and Bernstein's reporting on the Watergate break-in.} exposing bad actors or promoting a just cause \cite{jackson2020hashtagactivism}.

In this work we follow the \emph{working definition} provided by \citet{douglas2023conspiracy}: ``A conspiracy theory is a belief that two or more actors have coordinated in secret to achieve an outcome and that their conspiracy is of public interest but not public knowledge.'' Conspiracy theories are (a) oppositional; (b) describe malevolent acts; (c) ascribe agency to small groups; (d) epistemically risky; and (e) shared in order to form new social realities.

%Posts on platforms such as Twitter/X are short, informal, and often highly dependent on implicit context. They may rely on sarcasm, rhetorical questions, coded language, or in-group slang. They may also refer to current events or prior conversations that are not explicitly mentioned in the post itself. Consequently, text that appears conspiratorial in isolation may turn out, with additional context, to be harsh political criticism, satire, or mockery of conspiracy believers. Conversely, seemingly vague or indirect posts may function as meaningful conspiracy signals within a specific community.
% Low resource and morphologically-rich languages such as Arabic and Hebrew add yet another layer of complexity to the task of algorithmic detection of conspiratorial discourse. The challenge stems not only from ambiguity introduced by the rich morphology, but also from the limited training over the language of a secluded community of speakers. \oren{not sure this is the right place for this. rethink!} 

In this work we propose an agentic approach for the detection of conspiratorial discourse. We first demonstrate the shortcomings of text-based approaches, then introduce our agentic framework: given a specific social media post, the agent can call tools to sample more content from the user, inspect the user's metadata, query the user's social circle, or utilize the structure of the social network.

% הורדתי כדי לקצר ומסבירים בהמשך בסקירת ספרות
% Previous work on misinformation and conspiracy detection has approached the problem in several complementary directions. Rule-based methods often rely on lexical indicators such as keywords, hashtags, or topic-specific phrases associated with known conspiracy narratives. Supervised machine-learning models have used textual, behavioral, and metadata-based features, while network-analysis approaches have examined how narratives spread, which users amplify them, and how communities form around shared claims or ideological positions. More recently, Large Language Models (LLMs) have been used as powerful text classifiers and reasoning-based detectors.

\begin{table*}[ht!]
\centering
\small
\setlength{\tabcolsep}{2pt}
\begin{tabular}{p{0.04\textwidth}p{0.66\textwidth}ccc}
\toprule
ID & Tweet text, English translation & Gold & GPT-5 & Gemini 3 Flash \\
\midrule
1 & Remember the crop-duster planes I photographed this afternoon?... Here is the result. Dirty and murky skies. This is what we breathe. This is the air pollution that is killing us. And by the way, they blame us for it. So wake up. & 1 & 1 & 1 \\

\addlinespace[3pt]

2 & A teaching assistant from Pardes Hanna tells her students during remote learning on “Zoom” that she does not know who killed former Prime Minister Yitzhak Rabin.
“It is not certain that Yigal Amir was the murderer, and there are many conspiracy theories surrounding the issue.”
The parents heard the lesson and filed a complaint with the principal, then with the Ministry of Education and the Pardes Hanna local council. & 0 & 1 & 1 \\

% \addlinespace[3pt]
% 3 & Many years ago, Benny Gantz was abducted by an alien spaceship during a secret operation beyond enemy lines. He was replaced by Robot-Gantz, whose mission is to destroy Israel and build an alien base on its ruins. All the details were found on Gantz’s phone, which is actually an intergalactic transmission device. The left is trying to hide this.
%  & 0 & 1 & 0 \\

\addlinespace[3pt]

3 & Like if you think the COVID vaccines contained strawberry-vanilla milkshake ingredients and tiny location-tracking chips, so the government could track us, including facial recognition technology, in order to put us all into a secret G5 database. Share this, because the media channels will not tell you the truth.
 & 0 & 1 & 0 \\

\addlinespace[3pt]
4 & I am suffering from severe pain, but my surgery has been
postponed. Hospitals in Jerusalem are striking because they lack
funding, while politicians dismiss the doctors' complaints as empty
slogans. This government is killing its people.
 & 0 & 0 & 1 \\

\addlinespace[3pt]

5 & People around the world are suing their governments to force them to fight global warming. It seems to us that we Israelis also have a strong case against one of the most indifferent governments in the West regarding the greatest problem facing our children and grandchildren.
 & 0 & 0 & 0 \\

\bottomrule
\end{tabular}
\caption{Examples illustrating context-dependent Hebrew conspiracy-tweet classification. Tweets are shown in English translation, and original Hebrew texts appear in Appendix~\ref{app:hebrew_examples}. Gold is the human label, where 1 indicates conspiracy endorsement and 0 indicates non-conspiracy. GPT-5 and Gemini 3 Flash are text-only predictions.}
\label{tab:motivating_examples}
\end{table*}

The shortcomings of text-based classification are illustrated in Table \ref{tab:motivating_examples} through five examples: both GPT-5 and Gemini 3 Flash agree on the correct prediction in examples 1\&5, agree on an incorrect prediction in example 2, and diverge over examples 3--4. 
Temporal context may be needed when a tweet indirectly refers to a recent event or news cycle. Conversational context may be needed when a tweet responds to, quotes, or mocks another post. Social context may be needed when interpretation depends on the author's usual stance, community, or repeated narratives. In such cases, the isolated tweet text may be insufficient to determine whether the author is endorsing a conspiracy narrative, criticizing it, joking about it, or merely reporting on it. In many such cases, the key distinction is therefore not whether conspiracy-related content is present, but whether the speaker is committed to it.

%\oren{this is a bit our of context. think how to better connect/move}
%Text-only classification remains limited when the meaning of a post depends on information outside the post itself. Temporal context may be needed when a tweet indirectly refers to a recent event or news cycle. Conversational context may be needed when a tweet responds to, quotes, or mocks another post. Social context may be needed when interpretation depends on the author's usual stance, community, or repeated narratives. In such cases, the isolated tweet text may be insufficient to determine whether the author is endorsing a conspiracy narrative, criticizing it, joking about it, or merely reporting on it. In many such cases, the key distinction is therefore not whether conspiracy-related content is present, but whether the speaker is committed to it. Table~\ref{tab:motivating_examples} illustrates this ambiguity using examples from the Hebrew conspiracy-candidate pool.

We argue that conspiracy-tweet classification should be treated as a contextual interpretation task rather than a purely textual classification task. Relevant context may come from several levels of the social-media environment: tweet-level metadata, the author's profile, the author's previous posts, and the interaction network surrounding the user. We treat these signals as contextual hints rather than as direct proof of user intent.
For example, consider the following tweet: 
\begin{quote}
    {\bf H1} \emph{``Those naive people don't realize the storm is part of Pfizer's secret agreement to take over the world, setting the New World Order. There's no hope for you.}''\footnote{Original Hebrew text is provided in Appendix~\ref{app:hebrew_examples}, Example~H1.}
\end{quote}

\noindent
The true label of the tweet is non-conspiracy. The multiple, very explicit conspiracy markers are used sarcastically, as the text suggests that a severe winter storm\footnote{``Winter Storm Carmel Brings One Month's Worth of Rainfall to Israel''. Haaretz, 12/22/2021. Accessed 9/14/2026.} is part of a plot involving the Pfizer vaccine.
Table~\ref{tab:context_example} shows how different workflows interpret the tweet. The text-only model mistakes strong conspiracy markers for endorsement, while contextual workflows recover the sarcastic reading using the storm context, the author's profile, prior weather-related posts, and ego-network evidence. The agents-debate workflow also reaches the correct label, reasoning about the absurdity and the parodic nature of the text.

\begin{table*}[t]
\centering
\small
\setlength{\tabcolsep}{2pt}
\begin{tabular}{p{0.13\textwidth}cp{0.30\textwidth}p{0.44\textwidth}}
\toprule
Setting & Pred. & Retrieved context excerpt & Main agent reasoning \\
\midrule

Text only & 1 &
H1. No external context. &
``The tweet explicitly references classic conspiracy tropes, including a `secret agreement' by Pfizer to `take over the world' and the establishment of a `New World Order'.'' \\

\addlinespace[3pt]

Tweet + author context & 0 &
Author profile: the user describes herself as liking history, singing, and ``talking nonsense.'' Tweet metadata: posted on Dec. 20, 2021. &
``The user's profile explicitly states they like to `talk nonsense,' further supporting the interpretation of the tweet as a joke or mockery rather than a genuine conspiracy theory.'' \\

\addlinespace[3pt]

Previous tweets & 0 &
Recent posts include ``How is the weather outside?'' and ``I bought pajamas for Carmel,'' shortly before the target tweet. &
``User history confirms they were discussing the weather and personal preparations for the storm, indicating the tweet is a joke rather than a genuine conspiracy belief.'' \\

\addlinespace[3pt]

Network context & 0 &
Ego-network users mention Storm Carmel on the date of the tweet. The ego network also includes mainstream news or political accounts and ordinary Israeli Twitter users. &
``Storm `Carmel,' which occurred on the date of the tweet, was confirmed by ego-context mentions of the storm by @user1 and @user2. [...] The ego-context includes mainstream journalists and regular users, suggesting general Israeli Twitter discourse where this brand of political sarcasm is common.'' \\

\addlinespace[3pt]

Two-agent debate & 0 &
No external retrieval. Analyst 2 argues for sarcasm. &
``Analyst 2's argument regarding sarcasm is convincing. The absurdity of linking a natural storm to a Pfizer `secret agreement' for a `New World Order' indicates parody.'' \\

\bottomrule
\end{tabular}
\caption{Changes in agent reasoning given different contexts to the same text (H1). Labels follow the binary annotation scheme: 1 indicates conspiracy endorsement and 0 indicates non-conspiracy. Each setting combines the original tweet with only the context specified in that row.}
\label{tab:context_example}
\end{table*}

However, simply appending all available contexts to every tweet, as part of the input, is neither efficient nor necessarily desirable. 
%Different tweets require different forms of evidence: some can be classified from text alone, whereas others require metadata, author history, or network context. 
%While Table \ref{tab:context_example} demonstrates the necessity of contexts, we argue that contexts should be introduced selectively. 
Providing excessive contexts may introduce noise and harm performance \cite{shi2023large}. Agentic workflows are, therefore, suitable for this setting as an agent can reason and decide what tool should be used next, or whether the data available is sufficient.

We evaluate our agentic approach on a unique and challenging dataset of Hebrew tweets. The agentic approach achieves the best results compared with both the text-only LLM baseline and the corresponding preloaded-context conditions, improving F1 from 0.535 to 0.730 over the text-only LLM. We further experiment with a ``Socratic'' two-agent setting, in which one agent questions the reasoning of another agent, encouraging critical reflection. Finally, we provide an analysis of the tool-use behavior.
%use of tools, the frequency of the call for each tool and the order in which tools are evoked. 

Specifically, we focus on the Israeli X (formerly Twitter) ecosystem, leveraging a unique dataset estimated to cover 80--90\% of public Hebrew tweets from December 2018 to June 2023. 
Low-resource and morphologically rich languages such as Arabic and Hebrew add yet another layer of complexity to the task of algorithmic detection of conspiratorial discourse. The challenge stems not only from ambiguity introduced by the rich morphology, but also from the limited training over the language of a secluded community of speakers.
The dataset spans multiple waves of a global pandemic as well as five general-election cycles reflecting intense political tensions, a fertile ground for contentious narratives, conspiratorial thinking, and increased polarization.
Together, these social and linguistic conditions make Hebrew conspiracy-tweet classification a particularly challenging setting for automatic detection.

% In addition to retrieval-based context, we examine heterogeneous multi-agent debate as a complementary mechanism for handling interpretive ambiguity. In many cases, the difficulty is not only that information is missing, but that the same evidence admits multiple plausible readings. Different LLMs may notice different rhetorical cues, assign different weights to contextual signals, or reach different conclusions about whether a tweet endorses or mocks a conspiracy narrative. A multi-agent debate can expose competing interpretations and encourage reconsideration before the final classification decision \cite{liang2024encouraging}.

%As a secondary experiment, we also test whether heterogeneous two-agent debate improves text-only classification through critique and reconsideration \cite{liang2024encouraging}.

\paragraph{Contribution} Evaluating our context-aware agentic retrieval as well as heterogeneous two-agent ``debate'' framework, the contribution of this paper is threefold: (i) We demonstrate that the use of various contexts significantly improves the detection of conspiratorial posts, (ii) We explore the patterns of efficient tool-usage, and (iii) We analyze the token-economy tradeoffs associated with the agentic framework.

% \begin{itemize}
%     \item \textbf{RQ1:} Does contextual information improve conspiracy-tweet classification beyond isolated tweet text?
%     \item \textbf{RQ2:} Which forms of context, individually or in combination, are most informative for classification?
%     \item \textbf{RQ3:} Does agent-driven retrieval offer advantages over preloaded-context prompting?
%     % \item \textbf{RQ4:} Can heterogeneous multi-agent debate improve text-only conspiracy classification without external context? \oren{not sure this last one should be a RQ, given results are not great and we didn't investigate it properly.}
% \end{itemize}

\section{Related Work}
\label{sec:related}

\paragraph{Conspiracy Theories and Online Communities}

The World Economic Forum's Global Risks Report 2026 \cite{wef2026global} lists misinformation and disinformation among the ``most severe global risks''.

False narratives are often emotionally engaging and more appealing than accurate information \cite{vosoughi2018spread}, making them an effective tool for garnering attention and promoting political agendas \cite{starbird2017examining,starbird2019disinformation}. Conspiracy theories are a prominent form of this broader information disorder, explaining complex societal, political, medical, or global events as the result of secret plots by powerful actors. The appeal of conspiratorial claims is associated with the needs for certainty, control, identity, and belonging \cite{douglas2017psychology,douglas2019understanding}. Examples include claims that vaccines cause autism or that climate change is a deliberate fraud \cite{delvicario2016spreading}. In Israel, a 2015 survey found that 19\% questioned the identity of Prime Minister Yitzhak Rabin's assassin, while 35\% questioned some part of the official account \cite{caspit2015rabin}.

Online platforms serve as the vehicle through which various narratives are shared and reinforced. Existing network structures are harnessed and new communities are created and maintained by committed individuals. These communities provide a sense of belonging, promoting bespoke interpretations of reality \cite{delvicario2016spreading,douglas2017psychology}. 

This ``asymmetry of passion'' \cite{diresta2024invisible} allows small but highly committed online groups to amplify fringe narratives, creating an illusion of consensus and achieving offline impact \cite{vosoughi2018spread,starbird2019disinformation}.

\paragraph{Conspiracy Detection}

The body of computational work explicitly addressing the detection of conspiracy theories is surprisingly sparse. This subsection provides a brief survey of the literature as well as relevant works concerned with related tasks such as stance detection and the classification of `fake news' and misinformation.

Treating conspiracies as narrative structures, rather than true/false statements, \citet{shahsavari2020conspiracy} combine semantic role labeling, BERT embeddings, clustering, and community detection to uncover the narratives in emerging COVID-19 conspiracy theories. 
Similarly, \citet{lei2023identifying} enhance a transformer-based classifier with event-relation graphs to capture narrative structure in long-form news, showing that such structural information improves conspiracy detection and generalization to unseen media sources.
Beyond purely textual representations, \citet{steffen2025more} use unsupervised multimodal topic modeling to analyze textual and visual conspiracy discourse in German-language Telegram channels.

A number of datasets related to various conspiracy theories were also made available. COCO provides tweets covering twelve types of COVID-related conspiracies, together with benchmarks for stance and topic classification \cite{langguth2023coco}. YouNICon extends such resources to YouTube, providing a large-scale collection of conspiracy-related videos with manually labeled subsets for conspiracy detection and topic classification \cite{liaw2023younicon}. 
Related work has also evaluated transformer-based models on topic-diverse conspiracy data. \citet{phillips2022hoaxes} introduce a Twitter dataset that spans climate change, COVID-19 origins and vaccines, and the Epstein-Maxwell trial, and evaluate models of the BERT-family for conspiracy, stance, and topic classification.
Based on this data set, \citet{george2024conspiracy} augment a RoBERTa-based conspiracy classifier with text-derived psycho-linguistic features that capture emotion and moral framing, showing that these additional signals improve classification over text alone.
Extending this line of work beyond topic-specific classification, \citet{fort2023bigfoot} examine whether models trained on one conspiracy topic can generalize to unseen topics. They find that BART outperforms SVM baselines, but cross-topic performance remains variable and can decline substantially.

\citet{diab2024conspiratorial} further show that discussing a conspiracy does not necessarily imply endorsing it, using human-labeled Reddit posts to train BERT-family classifiers and evaluate zero- and few-shot GPT prompting for distinguishing conspiracy endorsement from criticism, skepticism, and debunking.

A few works use LLMs for conspiracy-related tasks. LLMs were found to perform similarly to fine-tuned transformers in detecting conspiracies in German Telegram data \cite{pustet2024detection}. 
\citet{liu2024conspemollm} instruction-tune an emotion-oriented LLM for conspiracy detection, topic classification, and stance-related tasks, showing improvements over general-purpose LLMs and ChatGPT on most evaluated tasks.
Complementarily, \citet{corso2025conspiracy} evaluate open-weights LLMs in a zero-shot setting for detecting conspiracy theories from TikTok video transcripts, finding that some prompting configurations achieve high precision, outperforming a fine-tuned RoBERTa baseline, although performance varies substantially across models and prompts.

\paragraph{Contextual Detection of Misinformation} Conspiracy detection is closely related to the broader literature on misinformation, rumors, and fake-news detection, where additional contextual and social signals have been incorporated to facilitate accurate classification.

At the content level, approaches have combined lexical, topic, hashtag, and URL features with information from the source message \cite{castillo2011information,qazvinian2011rumor}. 
User- and source-level signals have also been incorporated, including user credibility and stance \cite{li2019rumor}, as well as source-user behavior jointly modeled with article content \cite{ruchansky2017csi}. 
Temporal context has been captured from sequences of related posts and user responses, modeling how rumor discussions evolve over time \cite{ma2016detecting}, while propagation and network structure have been represented through diffusion features \cite{castillo2011information} and graph-based approaches such as Bi-GCN, which models both top-down propagation from the source post and bottom-up dispersion through the discussion graph \cite{bian2020rumor}.

Together, these works show that user's stance, and engagement patterns, along with propagation structure, provide important signals that could be harnessed to achieve improved classification.

\paragraph{Agentic Frameworks for Fact Verification}

Agentic approaches have been applied to evidence gathering and verification in misinformation detection. FactAgent decomposes fake-news verification into a structured expert workflow that combines the LLM's internal analysis with external search and source-credibility tools \cite{li2024large}. MARO assigns subtasks such as linguistic analysis, external fact consistency, and user-comment analysis to specialized agents, and automatically optimizes the decision rules used to combine their analysis for cross-domain misinformation detection \cite{li2025maro}. 
An agentic LLM approach for conspiracy-endorsement detection was designed to mitigate the ``Reporter Trap,'' in which objective reporting is incorrectly classified as endorsement, on the PsyCoMark benchmark \cite{spanakis2026agentic}.

We note that while the tasks of fact verification and the detection of misinformation bear similarities to the task at hand, they also differ significantly. We are not interested in the `truth' value of a statement but in its pragmatic function.\footnote{Consider example H1 above. The factually wrong content is used as a rhetorical device, mocking and rejecting the conspiracy.}

While our work is inspired by these works, it leverages a distinct set of tools and evidence sources appropriate for this interpretive setting. Rather than retrieving factual evidence to verify a claim, or applying a fixed set of analytical dimensions, our agent can selectively retrieve tweet metadata, the author's profile, prior discourse, and local ego-network context according to the needs of each case. This also distinguishes our approach from recent agentic conspiracy-detection work, which reasons primarily over the target text, retrieved examples, and limited contextual signals, without incorporating user history or broader discourse and network context. Our approach is also unsupervised, requiring no task-specific fine-tuning. 

%We evaluate this framework on a broad, multi-topic Hebrew conspiracy dataset and directly compare autonomous retrieval with matched conditions in which the same contextual sources are supplied in advance, allowing us to isolate the contribution of agent-controlled evidence acquisition. 

To the best of our knowledge, this is the first work to apply an agentic framework to conspiracy detection using tools that selectively retrieve user information, prior user discourse, and network context.

\section{Data}
\label{sec:data}
%Our data pipeline stems from high-recall candidate retrieval over a large Hebrew Twitter archive and culminates in the construction and human annotation of a targeted 504-tweet challenge set.

\subsection{Data Curation}
\label{subsec:candidate_retrieval}

\paragraph{Raw Data} We used a large archive of Hebrew-language tweets collected between December 2018 and June 2023. The archive, containing 270 million tweets posted by 3.3 million unique users, is estimated to cover 80--90\% of public Hebrew Twitter during this period.\footnote{We used Twitter's streaming API, tracking a list of Hebrew stopwords. The 80\%--90\% coverage is estimated by extrapolating the coverage achieved by the tracked terms over Hebrew tweets collected using the Decahose, which provided a random sample of 10\% of the public stream.} %For candidate construction, we retain only original tweets, focusing on authored content rather than retweeted material.

\paragraph{Conspiracy-related Tweets} We followed previous work on the creation of conspiracy datasets and used a comprehensive set of keywords in order to extract conspiracy-related content. This stage is designed to achieve ``high recall'', resulting in an adversarial dataset, retaining tweets that either contain conspiracy narratives or discuss related topics (e.g., questions and legitimate concerns regarding vaccines as well as anti-vaccination content).
The retrieval rules combine topic-specific keywords and general conspiracy cues. Topics include flat-earth claims, the Rabin assassination, the Yemenite children affair, COVID, medical and pharmaceutical control, global elites and the Illuminati, QAnon, chemtrails, 9/11, climate change, the moon-landing hoax, UFOs and aliens, and the ``deep state''. Brief descriptions of these topics, together with the full retrieval lexicon, are provided in Appendix~\ref{app:key_terms}.

A tweet should match more than one filter in order to be considered. For example the following tweet, retrieved by the chemtrails-related rules.

\begin{quote}
{\bf T1} \emph{``Remember the photo of the crop-duster planes I took this afternoon? Here is the result. Dirty and murky skies. This is what we breathe. This is the air pollution that's killing us. And by the way, they blame it on us. So wake up.''}\footnote{Original Hebrew is available in Appendix~\ref{app:hebrew_examples}, Example~T1.}
\end{quote}

\noindent
The tweet matches several topic-related terms and phrases, including ``air pollution'', ``plane'', ``spraying''\footnote{In Hebrew, the term used for a `crop-duster plane' is a `spraying plane', after its aerial application of fertilizers and pesticides.}, and ``sky'', together with broader conspiracy cues such as ``killing us'' and ``wake up''. The example illustrates the purpose of the retrieval stage: the rules identify tweets containing topic-related and potentially conspiratorial signals, but do not determine the label. The full list of keywords, and retrieval rules and thresholds can be found in Appendix \ref{app:key_terms}.
Removing duplications, this process produced 33,121 unique tweets, each likely to relate to one of the topics listed above.

\paragraph{Adversarial Subset} We first used Gemini 3 Flash and GPT-5 to independently classify all 33,121 candidates based on text alone, using the preliminary sampling prompt provided in Appendix~\ref{app:sampling_prompt}. 
We used disagreement between Gemini 3 Flash and GPT-5 as a signal for cases in which text-only predictions differed across models. The two models disagreed on 3,915 of the 33,121 candidates (11.8\%).

A subsample of 504 tweets was annotated by two annotators, see details in Section \ref{subsec:annotation} below. The annotated dataset is used to evaluate our classification pipeline. Once validated, the pipeline can be applied to the larger adversarial set. The raw dataset is available to the agents to query, using the various tools. 

\subsection{Data Annotation}
\label{subsec:annotation}
We sampled a set of 504 tweets for manual annotation. Cases in which Gemini 3 Flash and GPT-5 produced conflicting predictions were over-sampled. The distribution of tweets by models agreement groups is available in Table~\ref{tab:eval_composition}. 

% \begin{table}[t]
% \centering
% \small
% \begin{tabular}{ccr}
% \hline
% Gemini 3 Flash & GPT-5 & Count (\%) \\
% \hline
% 0 & 0 & 26 (5.2\%) \\
% 0 & 1 & 234 (46.4\%) \\
% 1 & 0 & 212 (42.1\%) \\
% 1 & 1 & 32 (6.3\%) \\
% \hline
% \multicolumn{2}{l}{Total} & 504 (100\%) \\
% \hline
% \end{tabular}
% \caption{Composition of the evaluation set according to text-only predictions from Gemini 3 Flash and GPT-5.\oren{Lior - can you add another column with the distribution over the 33,121 tweets?}}
% \label{tab:eval_composition}
% \end{table}

\begin{table}[t]
\centering
\small
\begin{tabular}{ccrr}
\hline
Gemini 3 Flash & GPT-5 & Annotated & Candidate pool \\
 & & Count (\%) & Count (\%) \\
\hline
0 & 0 & 26 (5.2\%) & 19,283 (58.2\%) \\
0 & 1 & 234 (46.4\%) & 2,835 (8.6\%) \\
1 & 0 & 212 (42.1\%) & 1,080 (3.3\%) \\
1 & 1 & 32 (6.3\%) & 9,923 (30.0\%) \\
\hline
\multicolumn{2}{l}{Total} & 504 (100\%) & 33,121 (100\%) \\
\hline
\end{tabular}
\caption{Composition of the evaluation set and the full candidate pool according to text-only predictions from Gemini 3 Flash and GPT-5. 0(1) indicates a prediction of not (conspiracy). Note the balanced-oversampling of `0 1' and `1 0'.}
\label{tab:eval_composition}
\end{table}

Manual annotation of conspiracy-related tweets is challenging and time consuming. The annotator is required to have a deep understanding of the different topics and related conspiracies. These involve some bizarre claims and in-group lingo, internet culture, and niche use of memes. Labels should be assigned based on the communicative function of the tweet rather than its surface wording alone, and the working definition provided by \citet{douglas2023conspiracy}. The adversarial construction of the dataset (see above) makes the annotation even more challenging. Informed decision is often made only after careful examination of the different contexts -- a time consuming and cognitively draining undertaking. 

Due to the complexity of the annotation process we took the following approach: All 504 tweets were manually labeled by two domain experts. The first annotated each tweet from scratch: model predictions were hidden and all relevant contexts were examined in order to assign the gold label. The second expert reviewed the labels assigned by the first annotator, exploring the contexts only in cases the original label seemed lacking without further evidence. In the final stage the annotators had a consolidation session, reaching agreement on the gold labels, resulting in 331 non-conspiracy tweets (65.7\%) and 173 conspiracy tweets (34.3\%). %The annotated dataset is available at [URL]\footnote{Will be added upon acceptance.}.

%Annotation began with the target tweet itself, but when its intended meaning could not be determined reliably from the text alone, the annotator could consult the same categories of contextual evidence later made available to the agentic workflows, including author information, previous tweets, and social-network context. This context-aware procedure was intended to assign labels according to the communicative meaning of the tweet rather than its surface wording alone.

\section{Methods}
\label{sec:methodology}

\subsection{The Agentic Pipeline}
Agentic frameworks extend language models beyond direct generation by enabling them to interact with external tools and environments during inference. ReAct \cite{yao2023react} exemplifies this by interleaving reasoning with actions and observations in an iterative loop with an external environment, while Toolformer \cite{schick2023toolformer} takes a more static approach, learning when and how to invoke external APIs and incorporate their outputs into generation without an ongoing interactive loop.

In this work we adopt the former approach. The agent is prompted with the task definition and specific guidelines introducing the available tools, regulating the tool use, and defining the input-output formats used through the reasoning and acting loop. This agentic workflow is illustrated in  Figure~\ref{fig:fig_agent_tool_workflow} and the full prompt is available in Appendix \ref{subapp:full_agent}.

\begin{figure}[th!]
\centering
\includegraphics[width=\columnwidth]{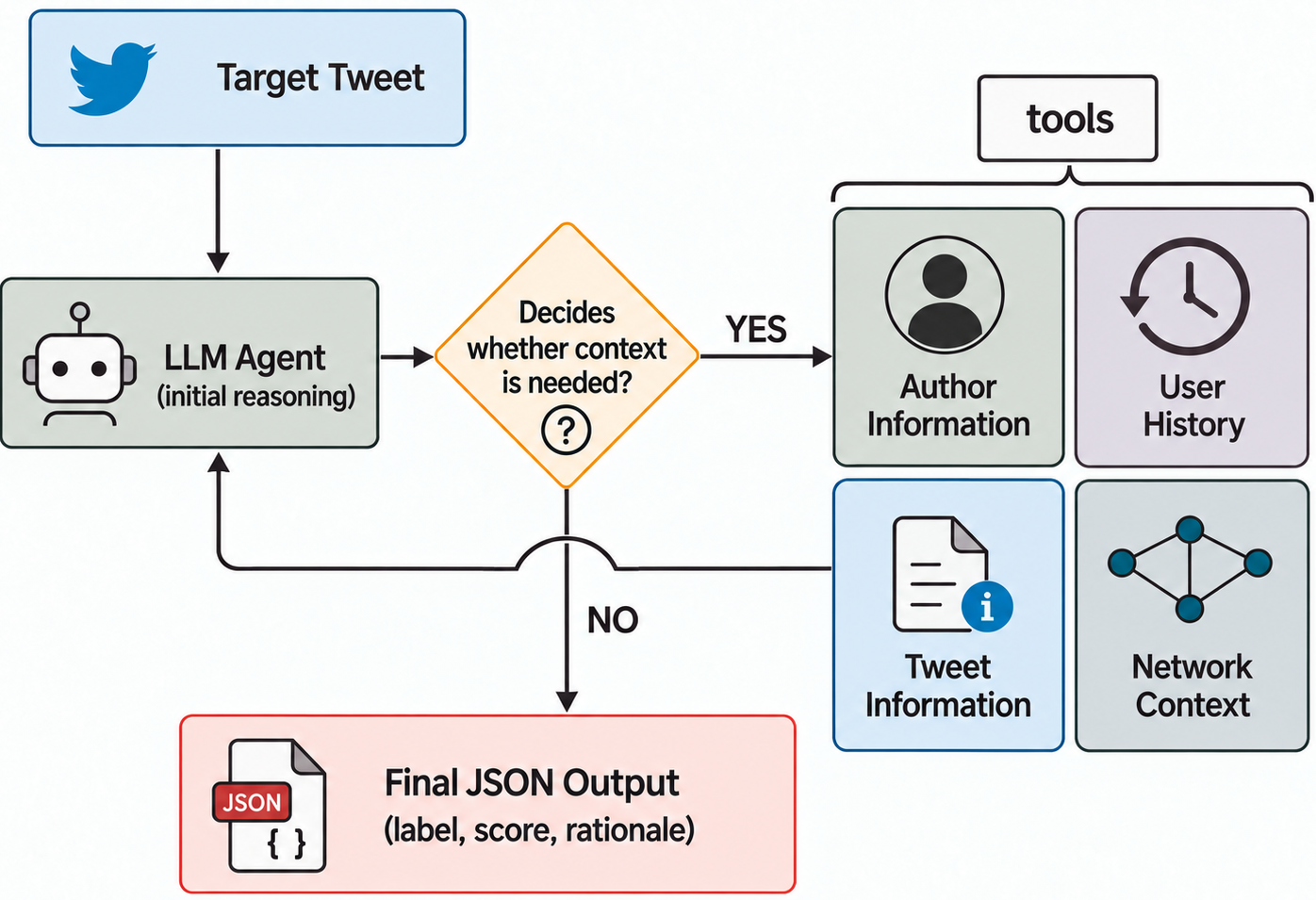}
\caption{Full agentic workflow.}
\label{fig:fig_agent_tool_workflow}
\end{figure}

\subsection{Contexts and Tools}
\label{subsec:context_sources}

Prior work, mostly on fake-news and misinformation, has demonstrated the utility of social context for classification tasks. Fortunately, our unique \emph{raw} dataset provides a good approximation of the full knowledge of users' public history and social interactions. A set of four tools was developed, allowing the agent to query and retrieve different types of context, as the need arises in the reasoning loop. 
We note that tools only serve data that was timestamped \emph{before} the target tweet was posted, masking any ``futuristic'' signals. The available tools are described in the remainder of this Subsection.

\subsubsection{Tweet Information (metadata)}
\label{subsubsec:tweet_information}

The tweet-information tool retrieves metadata about the target tweet, including its timestamp, text length, hashtags, URLs, and source. These signals can situate the tweet in a temporal or topical context and indicate links to external material or unusual posting behavior.

\subsubsection{Author Information (metadata)}
\label{subsubsec:author_information}

The author-information tool retrieves profile-level information about the tweet author, including username, display name, bio, account creation date, verification status, follower and following counts, total tweet count, language, and location when available. Such information can provide cues about the author's identity, self-presentation, or broader account characteristics that may help interpret intent.

\subsubsection{User History}
\label{subsubsec:user_history}
This tool samples up to 20 tweets posted by the same user in the 30 days preceding the target tweet. This context can reveal prior stance, recurring narratives, repeated references, or ongoing themes that may clarify whether the target tweet endorses, mocks, or merely references a conspiracy narrative.

\subsubsection{Network-Aware Ego Context}
\label{subsubsec:network_context}

The network tool recovers and returns the ego-network of the user, based on retweet interactions occurring in the 14 days preceding the target tweet. Nodes are users whose content the target author retweeted during this period. This signal provides the immediate community of the user and a glimpse into the content she recently amplified. Nodes are ranked by interaction frequency, and the five most frequently retweeted accounts are retained. For each of these nodes, the tool returns basic profile information and up to five original tweets posted during the 30 days preceding the target tweet.

The restrictions on the number of tweets returned (20 or 5, depending on the tool), the timespan from which we sample (14  or 30 days) or the number of neighbors considered, are applied for efficiency purposes, curbing down runtime. Our purpose here is to demonstrate the advantage of the agentic framework using the specified tools, thus optimizing these hyperparameters is beyond the scope of this work.

\subsection{Baseline Models}
%\oren{consider removing all references to Dicta from text and tables. the results are based on a different annotation protocol.}
The following settings are used as baselines: 

\subsubsection{Tweet-only LLM} In the most basic setting, an LLM (Gemini 3 Flash) is prompted with the target tweet and the task definition. No contexts are provided, and only the tweet is considered.

\subsubsection{Degenerate Agent} A Gemini 3 Flash model stripped of its tools sets the baseline in our experiments. In this setting, the context(s) are preloaded and appended to the target tweet and the classification instructions in the original prompt. The degenerate agent does reason about the provided contexts, but it has no agency as it cannot actively decide on tool use through the reasoning loop.

% \subsubsection{A Fine-tuned Supervised Model} DictaBERT is a BERT-based model modified and trained to account for the rich morphology of Hebrew \cite{shmidman2023dictabertstateoftheartbertsuite}. 
% We fine-tuned DictaBERT on a validation set. After excluding the fixed 504-tweet evaluation set and removing repeated texts, 742 tweets remained. Using random seed 42, we divided them with a stratified 80/20 split into 593 training and 149 validation tweets. There is no overlap in tweet IDs or tweet texts across the training, validation, and evaluation sets. DictaBERT was fine-tuned for four epochs with a learning rate of $3\times10^{-5}$ and batch size 8, and the checkpoint with the highest validation F1 was selected for evaluation.

\begin{figure}[h!]
    \centering
    \includegraphics[width=0.9\columnwidth]{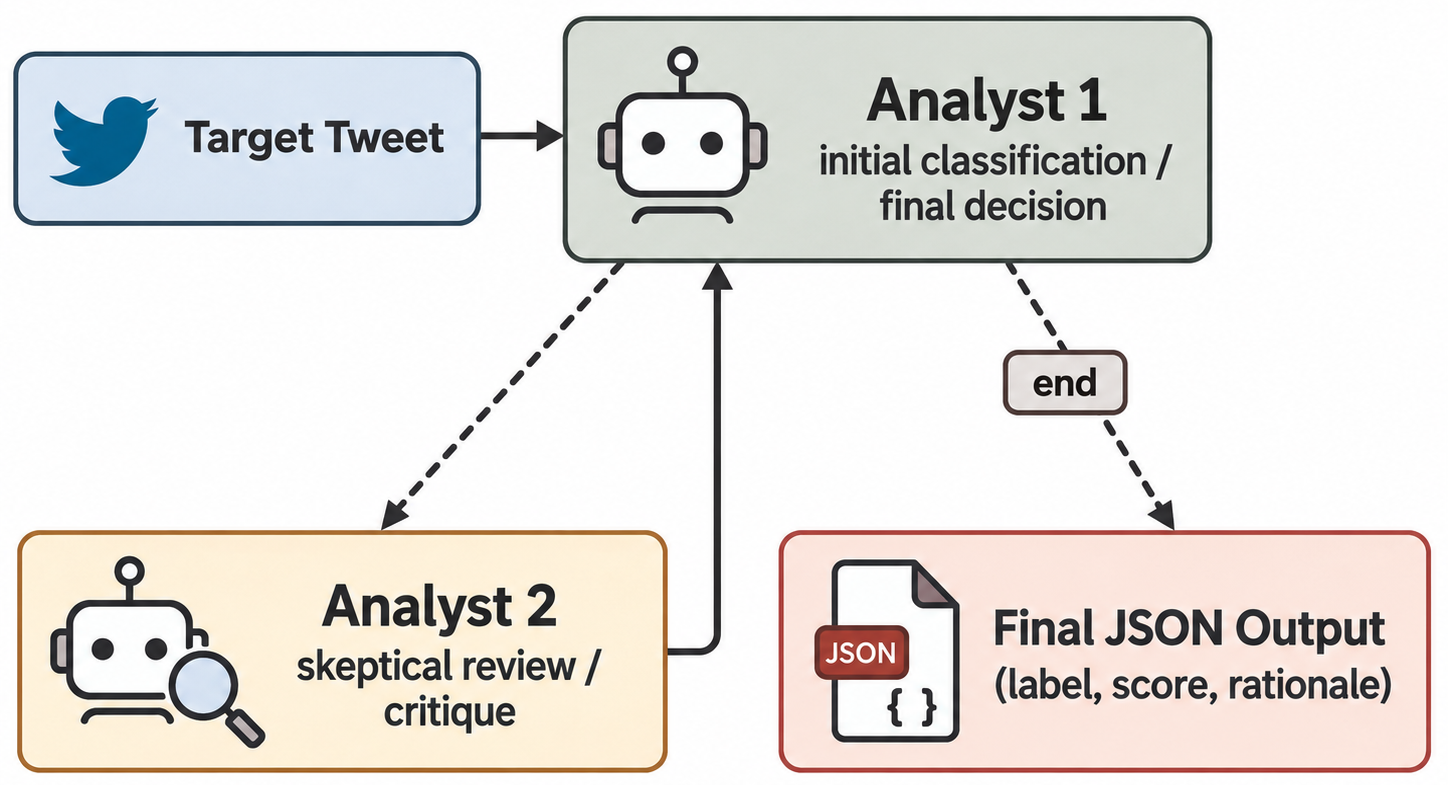}
    \caption{Two-agent debate workflow. Analyst 2 critiques Analyst 1's initial decision before the final prediction.}
    \label{fig:fig_debate_workflow}
\end{figure}

\subsubsection{Agents-debate Baseline}
\label{subsec:debate_method}
Multi-agent debates provide a complementary mechanism for exposing alternative interpretations through critique and reconsideration \cite{liang2024encouraging,han2025debate,liu2025truth}. We use a simple debate setting in which one model (Analyst 1) classifies a target tweet, then invites another model (Analyst 2) to debate and criticize its reasoning and classification decisions. Analyst 2 is instructed to pay particular attention to possible non-literal interpretations. 
Given the critical response provided by Analyst 2, Analyst 1 re-weighs her position and produces the final decision. Each analyst is only exposed to the target tweet and to the reasoning and decisions made by the other analyst. This workflow is illustrated in  Figure~\ref{fig:fig_debate_workflow} and the respective prompts and protocol are available in Appendix \ref{subapp:debate}. Note that the analysts differ both in the assigned persona and in the foundation model used -- Gemini 3 Flash and GPT-5 as Analysts 1 and 2, respectively.

% \subsection{Formal Task Definition}
% \label{subsec:task_definition}

\subsection{Experimental Setting}
\label{subsec:exp_design}

We formulate the task as binary tweet-level classification. Given a target Hebrew tweet, the system predicts whether it should be labeled as conspiracy (1) or non-conspiracy (0). In all settings, the model (agent) returns a structured output containing the predicted label along with a short rationale (see Table \ref{tab:context_example}). 
We report accuracy, precision, recall and F-Score over the annotated dataset, establishing the basic performance of each model and setting. 

Our experimental settings, described below, aim to separate three factors that are often conflated in context-aware classification: (i) the availability of contextual evidence; (ii) the type of evidence provided; and (iii) the mechanism through which that evidence is accessed. 

We thus have two main experimental settings: the \emph{Preloaded} mode, using degenerate agents, and the \emph{Agentic mode}. In the Preloaded mode the agent receive the context by default, whereas in the Agentic mode agents determine whether to access a context and when.

Within each of these settings, we assess the contribution of each context/tool separately and in a full setting in which all contexts/tools are available to the agent. The various settings are further explained in Table \ref{tab:experimental_conditions}.

\begin{table*}[th]
\centering
\small
\setlength{\tabcolsep}{4pt}
% \begin{tabular}{@{}>{\raggedright\arraybackslash}p{0.18\columnwidth}
%                 >{\raggedright\arraybackslash}p{0.21\columnwidth}
%                >{\raggedright\arraybackslash}p{0.58\columnwidth}@{}}
\begin{tabular}{lcl}
\hline
{\bf Setting} & {\bf Context / Tools}& {\bf Description} \\
\hline
LLM & -- & Gemini 3 Flash classifies the tweet without added context. \\
\hdashline
\multirow{4}{*}{Degenerate}
 & Tweet+User & Tweet and author information are inserted into the prompt. \\
 & History & User history is inserted into the prompt. \\
 & Network & Ego-network context is inserted into the prompt. \\
 & ALL & All context sources are inserted into the prompt. \\
\hdashline
Debate & -- & Gemini 3 Flash and GPT-5 classify through the fixed debate workflow. \\
\hdashline
\multirow{4}{*}{Agentic}
 & Tweet+User & Agent may retrieve tweet and author information. \\
 & History & Agent may retrieve previous tweets by the same author. \\
 & Network & Agent may retrieve ego-network context. \\
 & ALL & Agent may use all contextual tools. \\
\hline
\end{tabular}
\caption{Experimental conditions evaluated in the study.}
\label{tab:experimental_conditions}
\end{table*}

\paragraph{Execution} The agentic workflows were implemented using LangChain \cite{langchain} and LangGraph \cite{langgraph}. Gemini 3 Flash was used for the text-only baseline and all single-agent LLM conditions. GPT-5 was used as the second agent in the debate setting. Temperature was set to 0.2 in all settings. Full prompts are provided in Appendix~\ref{app:prompts}.

% \subsection{Evaluation and Analysis}
% \label{subsec:evaluation}

\section{Results and Analysis}
\label{sec:results}
All settings (Table \ref{tab:experimental_conditions}) were first evaluated on the annotated dataset. We report overall accuracy and precision, recall and F1-score with respect to the positive class. Classification results complement with error analysis. 

We follow the classification results with analysis of the tool usage: exploring the utility, usage frequency, and temporal patterns in which tools are used. 

Finally, based on all reasoning steps, we explore the input/output tradeoff (in ``token economy'') between the agentic and the degenerate frameworks. 

\subsection{Overall Performance}
\label{subsec:overall_performance}

All classification results are presented in Table~\ref{tab:overall_results}. 
The baseline results are established by the Text-only LLM with accuracy of 0.6 and F1-score of 0.535. 
Best over-all performance (accuracy of 0.790 and F-score of 0.73) are achieved by the full agentic framework (all tools available). In comparison, the degenerate agent using all contexts achieves an accuracy of 0.707 and F-score of 0.67. 
Allowing only one tool, the best results are achieved with the User History context (Accuracy: 0.748; F: 0.698), outperforming the Degenerate Agent with full context. This last result demonstrates the benefits of using the agentic framework, compared to the use of elaborated contexts as input. 

The Two-agent Debate framework outperforms the text-only LLM baseline but falls short of all context-augmented LLM conditions in F1.

These results confirm, yet again, the challenge in conspiracy detection and the need for contexts, especially over our ``adversarial'' dataset.

\begin{table}[th!]
\centering
\small
\begin{tabular}{lcccc}
\hline
Setting & Acc. & Prec. & Recall & F1 \\
\hline
Text-only LLM & 0.603 & 0.447 & 0.665 & 0.535 \\
Two-agent Debate & 0.683 & 0.534 & 0.584 & 0.558 \\
Degenerate: Tweet+User & 0.691 & 0.540 & 0.669 & 0.597 \\
Degenerate: Network & 0.667 & 0.509 & 0.803 & 0.623 \\
Degenerate: History & 0.681 & 0.521 & 0.855 & 0.648 \\
Agentic: Tweet+User & 0.762 & \textbf{0.657} & 0.642 & 0.649 \\
Agentic: Network & 0.726 & 0.578 & 0.751 & 0.653 \\
Degenerate: Full Context & 0.707 & 0.546 & \textbf{0.866} & 0.670 \\
Agentic: History & 0.748 & 0.593 & 0.850 & 0.698 \\
Agentic: All tools & \textbf{0.790} & 0.653 & 0.827 & \textbf{0.730} \\
\hline
\end{tabular}
\caption{Overall classification performance across the evaluated conditions.}
\label{tab:overall_results}
\end{table}

We compare the text-only LLM with the best-performing setting
\textit{Agentic: All Tools} model, referred to below as the
full-context agent.

\begin{table}[th!]
\centering
\small
\setlength{\tabcolsep}{2pt}
\begin{minipage}{0.48\columnwidth}
\centering
\textbf{(a) Text-only LLM}\\[2pt]
\begin{tabular}{lcc}
\toprule
 & Pred.\ C & Pred.\ NC \\
\midrule
Gold C  & 115 & 58 \\
Gold NC & 142 & 189 \\
\bottomrule
\end{tabular}
\end{minipage}
\hfill
\begin{minipage}{0.48\columnwidth}
\centering
\textbf{(b) Agentic: All Tools}\\[2pt]
\begin{tabular}{lcc}
\toprule
 & Pred.\ C & Pred.\ NC \\
\midrule
Gold C  & 143 & 30 \\
Gold NC & 76 & 255 \\
\bottomrule
\end{tabular}
\end{minipage}

\caption{Confusion matrices for the text-only LLM and
\textit{Agentic: All Tools}. C denotes conspiracy and NC denotes
non-conspiracy.}
\label{tab:confusion_matrices}
\end{table}

Tables \ref{tab:confusion_matrices}(a) and (b) show that contextual retrieval improves classification across both classes. 
%Recall for conspiracy tweets increases from 66.5\% to 82.7\%, while recall for non-conspiracy tweets increases from 57.1\% to 77.0\%. 
In absolute terms, the full-context agent reduces false positives from 142 to 76 and false negatives from 58 to 30. Context therefore helps both
identify conspiratorial discourse and reject non-conspiratorial tweets
that contain similar surface-level language.

\subsection{Error Analysis}
\label{subsec:error_analysis}
In order to gain insight into the performance of the agent, we explore cases in which both the agent and the baseline erred as well as instances in which the baseline provided the correct classification while the agent misclassified it.

\paragraph{Errors introduced by context} Although achieving a significant overall improvement (F1 $0.535 \rightarrow 0.730$), in some cases, the agent (all tools available) misclassified tweets that were correctly classified by the baseline, 78.4\% of these are false positives. This result reminds us that sometimes words should be taken at face value and incomplete contexts, often depending on hyperparameter settings, may distort the interpretation.

\paragraph{Shared errors}

We qualitatively explored the cases over which both the baseline and the agent erred. We find that these cases fall under two main categories: (i) Failure to distinguish conspiracy claims from strong political or constitutional criticism and accusations of corruption. (ii) Tweets that explicitly mock a conspiracy theory or
discuss it without endorsing it. 
The following tweets, respectively, illustrate the two categories: (a) ``And who is responsible for the severe erosion of public trust in the justice system in recent years? Mainly the system's leaders: [...] dirty and inflammatory investigations [...], lies constantly and pursues selective enforcement [...]''; and (b) ``I checked with the Ministry of Pandemics. The new virus will arrive redacted – highly classified. [...]''  The original tweets with English translations are provided as Examples~E1 and~E2, respectively, in Appendix~\ref{app:error_examples}. 

We note, however, that the agent does classify many such cases correctly.

\subsection{Tool-Use Patterns}
\label{subsec:tool_use}

The trace logs show that an agent with tool access does not simply use all tools at her disposal, retrieving all contexts for every tweet. As evident from Table \ref{tab:tool_use_patterns} no tool was used by the agent in 10.9\% of the cases. 

Considering all runs in which the agent did not use any of the tools, both the agent and the text-only LLM achieved 94.5\% accuracy and produced identical predictions. In contrast, among the runs in which the agent requested at least one context to be retrieved, text-only accuracy was 56.1\%, compared with 77.1\%
for the agent, suggesting that contextual retrieval improves performance over the harder cases where text-only classification struggles.

Table \ref{tab:tool_use_patterns} also presents the patterns of consecutive tool usage. In 36.3\% of the cases the agent requested both the User Info and User History in the first step, making the final judgment immediately after. Three tools were used in about 25\% of the cases, with User Info + User History $\rightarrow$ Network being the most common pattern (12.3\% of the cases). 
The use-rate of all tools (regardless of call order) is presented in Table \ref{tab:full_context_tool_rates}. User Info (metadata) proved the most important tool, used in 89.1\% of cases. User History was called 81\% of the cases, while Tweet Info (metadata) and Network contexts were used in 21-25\% of the cases.

\begin{table}[t]
\centering
\small
\setlength{\tabcolsep}{2pt}
\begin{tabular}{@{}p{0.70\columnwidth}rr@{}}
\hline
Pattern & Count & Share \\
\hline
User Info + User History
    & 183 & 36.3\% \\

User Info + User History $\rightarrow$ Network
    & 62 & 12.3\% \\

No Tool Use
    & 55 & 10.9\% \\

User Info $\rightarrow$ User History
    & 47 & 9.3\% \\

Tweet Info + User Info $\rightarrow$ User History
    & 31 & 6.2\% \\

User Info $\rightarrow$ User History $\rightarrow$ Network
    & 28 & 5.6\% \\
\hdashline
Other Patterns
    & 98 & 19.4\% \\
\hline
\end{tabular}
\caption{Tool-use patterns in the On-demand Full Context condition.
``+'' marks tools called in the same step. ``$\rightarrow$'' marks a
later tool call after observing a previous result. Author Info, User
History, Tweet Info, and Network refer to the corresponding context
sources described in Section~\ref{subsec:context_sources}.}
\label{tab:tool_use_patterns}
\end{table}

\begin{table}[!ht]
\centering
\small
\begin{tabular}{lr}
\hline
Tool in On-demand Full Context & Run-use rate \\
\hline
Tweet information & 21.8\% \\
Network context & 24.4\% \\
User history & 81.0\% \\
User information & 89.1\% \\
\hline
\end{tabular}
\caption{Tool-use rates within the On-demand Full Context condition.}
\label{tab:full_context_tool_rates}
\end{table}

% \begin{table}[t]
% \centering
% \small
% \setlength{\tabcolsep}{2pt}
% \begin{tabular}{@{}p{0.70\columnwidth}rr@{}}
% \hline
% Pattern & Count & Share \\
% \hline
% Author Info + User History
%     & 183 & 36.3\% \\

% Author Info + User History $\rightarrow$ Network
%     & 62 & 12.3\% \\

% No Tool Use
%     & 55 & 10.9\% \\

% Author Info $\rightarrow$ User History
%     & 47 & 9.3\% \\

% Tweet Info + Author Info $\rightarrow$ User History
%     & 31 & 6.2\% \\

% Author Info $\rightarrow$ User History $\rightarrow$ Network
%     & 28 & 5.6\% \\

% Other Patterns
%     & 98 & 19.4\% \\
% \hline
% \end{tabular}
% \caption{Tool-use patterns in the On-demand Full Context condition.
% ``+'' marks tools called in the same step. ``$\rightarrow$'' marks a
% later tool call after observing a previous result. Author Info, User
% History, Tweet Info, and Network refer to the corresponding context
% sources described in Section~\ref{subsec:context_sources}.}
% \label{tab:tool_use_patterns}
% \end{table}

% The most common pattern combines author information and user history in the same step. Network context usually appears as an additional step after user-level context, rather than as the first source of
% evidence. This suggests that the agent often begins by inspecting the author and prior discourse, and turns to broader network context only when additional social evidence appears useful. The retrieval process also remains relatively short in most runs, with 87.5\% of tweets requiring at most two retrieval steps.

\subsection{Analysis of `Token Economy'} 
\label{subsec:tokeneconomy}

Token Economy is a colloquial term often used to describe, estimate, and manage the costs of LLM usage. Generally speaking, tokens are used in three distinct stages: input tokens in the prompt, generated tokens at the output, and reasoning tokens. Reasoning tokens may not be visible to the user and are often considered as ``hidden'' generated output. However, some models explicitly output the intermediate reasoning steps. API providers often charge per token, usually charging different rates for input and output tokens. 
Agentic frameworks introduce a tradeoff between input, reasoning and output tokens. While initial prompts are typically shorter, the reasoning loops and the retrieved information may result in a significant increase in the use of reasoning and (intermediate) output tokens \cite{bai2026ai}. 
These frameworks and billing policies directly translate tokens into dollar cost, creating real economic incentives to minimize token usage \cite{bergemann2025economics}.

While we do not have direct access to token usage,\footnote{Not all reasoning is available and specific token counts depend on the model's BPE and other optimization practices} we can provide a rough estimate through the number of initial and intermediate input and output character count. 

The best performing agent (all tools available) used an average of 11,434 input characters per instance, across all tool calls, compared with an average of only 8,251 characters in the degenerate agent setting with full context. The numbers of output characters produced (final and intermediate) by the models are 1048 (agent) and 614 (degenerate agent). Thus, the agent processed approximately $1.4\times$ as many input characters and produced approximately $1.7\times$ as many output characters. This non-negligible computational tradeoff should be considered alongside the improvement in F1-score (0.670 $\rightarrow$ 0.730) and accuracy (0.707 $\rightarrow$ 0.790), as well as changing model-pricing policies.
A more thorough analysis of this tradeoff is beyond the scope of this work. 

\section{Discussion}
\label{sec:discussion}

\paragraph{Contexts} The error patterns suggest that the central challenge in conspiracy-tweet
classification is not merely recognizing conspiratorial content, but
inferring the underlying function of the utterance, the illocutionary act, in terms of speech acts theory \citet{austin1975things}. These two  examples (original language in Appendix \ref{app:hebrew_examples}) illustrate the challenge and the benefits of the agentic framework:

\begin{quote}
    {\bf D1:} \emph{``My brother sent me a WhatsApp message about how lemons kill cancer cells and can treat several types of cancer, and how pharmaceutical companies hide it from us. And I'm the black sheep of the family.''}
   \\ \\
   {\bf D2:} \emph{``The Zombie Guide: The term `conspiracy theory' is already passé. From now on, when you encounter a truth you cannot or do not want to investigate, instead of saying `conspiracy theory,' say: the Earth is flat. That way, you can avoid thinking about and investigating the frightening truth''}
\end{quote}

The first tweet quotes a conspiratorial message verbatim, but uses the concluding words to mock it. Text-only LLMs tend to misinterpret this, as the `sheep' metaphor is common in conspiratorial discourse. The second example seems to mock conspiracy theories while in fact it promotes them by mocking its critiques. With contexts, however, the agents correctly classify the utterances. 
Context appears to help by calibrating the textual interpretation rather than merely adding more text. User history can reveal recurring stance,
rhetorical style, and narrative participation, while author and
ego-network information provide additional cues about the social and
discursive setting of a post. 

\paragraph{Multi-Agent Debate} Interestingly, the two analysts (agents) in the debate settings initially disagreed on 17.9\% of the target tweets. Following the critique offered by Analyst 2, Analyst 1 reversed her decision in 60\% of the cases. Of these changes, (66.7\%) corrected an initial error, while a correct decision was reversed in 33.3\% of the cases. While the debate's overall performance was unimpressive, this framework shows clear potential. This direction will be explored in future work.

\paragraph{Limitations} This work comes with some limitations. First, hyper parameters were not optimized. This work serves as a convincing proof of concept for the utility of the agentic framework on the specific task and over a challenging adversarial dataset. As such we did not optimize hyperparameters or explored other possible settings, e.g., using other LLMs in the different settings.
Consequently, the analysis of the results and the errors apply to this dataset and the specific models used. Other or future models may perform differently in terms of reasoning or token efficiency.

\paragraph{An Ethical Note} The use of user histories and network/social context should be attuned to privacy considerations, particularly when user-level information is retained, analyzed, or released. Moreover,  incorrectly labeling a user may incur social costs. The system is therefore better suited to research purposes rather than as an  automated moderation tool without human review. 

%Future work should also examine more systematically when context helps, when it misleads, and whether contextual retrieval can be combined effectively with multi-agent debate.

%The findings are specific to Hebrew Twitter, the evaluated model versions, and one run per tweet. Future work should test larger and more representative samples, additional languages and platforms, broader retrieval strategies,
%and robustness across repeated runs and model updates.

% The agentic workflow also introduces an efficiency trade-off. Although it exposes the model to less distinct contextual information, multiple tool calls increase total processed input. Context compression, pruning, and caching may reduce this overhead.

\section{Conclusion}
\label{sec:conclusion}

This paper examined Hebrew conspiracy-tweet classification as a context-dependent interpretation task. Using a large local archive of Hebrew
tweets and a targeted evaluation set of manually labeled tweets, we find that the full agentic approach performs better than other approaches, with statistically significant gains over the text-only LLM and the matched preloaded full-context condition, even when the same contextual information is available. 
Our findings suggest treating conspiracy detection on social media as a socially embedded interpretation problem rather than purely a
text-classification problem.

{\small
\bibliography{references}
}

\appendix

\section{Prompt Templates and Agent Instructions}
\label{app:prompts}

The following prompt templates and agent instructions document the LLM workflows used in this study. Angle-bracketed fields below denote values instantiated for each tweet. Experimental prompts are reproduced verbatim from the implementation, while the preliminary sampling prompt is provided in English translation.

\subsection{Preliminary Sampling Prompt}
\label{app:sampling_prompt}

The following prompt was used only to obtain preliminary text-only predictions for disagreement-based sampling of the evaluation set. Its outputs were not used as gold labels or as reported experimental results. The original prompt was written in Hebrew. An English translation is provided below.

\begin{lstlisting}
You classify Hebrew tweets according to their degree of conspiratorial content.

Important distinction: criticism, ridicule, or sarcasm toward conspiracy
theories should be classified as non_conspiracy.
Irony will often appear as exaggeration, humorous tone, or ridicule without
promoting a secret plot.
If the author describes conspiracy theories in order to refute or ridicule
them, classify the tweet as non_conspiracy even if conspiracy-related content
is mentioned.

Task:
Classify each tweet at two levels:
1. Label: whether the text is conspiratorial
   ("conspiracy" / "non_conspiracy")
2. Score: degree of conspiratorial content between 0 and 1

In addition, characterize the text according to the following criteria
(one or more may apply):
1. A claim involving a secret connection or hidden plot
2. Contradiction of an accepted official explanation or position
3. Use of unreliable sources or vague language
   ("they say...", "everyone knows...")
4. Encouragement of distrust toward major institutions
   (government, science, media, etc.)
5. An "us versus them" framing

Input:
You will receive a list of tweets. Each tweet has a row_id and text.

Required output:
Return only valid JSON as a list of objects, in the same order and with
the same number of items as the input.

For each item, return:
{
  "row_id": "the received identifier",
  "label": "conspiracy" or "non_conspiracy",
  "score": number between 0 and 1,
  "signals": [list of numbers from {1,2,3,4,5}],
  "rationale": "two short sentences explaining the classification"
}

Rules:
- Keep the response short and clear.
- Do not add any text outside the JSON.
- Make sure to distinguish conspiratorial claims from irony or ridicule.
\end{lstlisting}

\subsection{Common Tweet Input}

The following user-message template was used for single-agent classification:

\begin{lstlisting}
Classify this tweet.
tweet id: <tweet_id>
day: <day>
tweet text: <tweet_text>
\end{lstlisting}

\subsection{Text-Only Classification}

\begin{lstlisting}
You classify whether a Hebrew tweet is conspiracy or non_conspiracy.

Classify from the tweet text alone.

Return only valid JSON:
{
  "label": 0 or 1,
  "score": number between 0 and 1,
  "rationale": short explanation grounded in the tweet text
}

Rules:
- score - conspiracy confidence (0 = not conspiracy, 1 = conspiracy)
\end{lstlisting}

\subsection{On-Demand Retrieval}
\label{subapp:full_agent}

The on-demand conditions used the same prompt structure, with only the
descriptions of the tools available in a given condition included. The
\textit{On-demand Full Context} system prompt is reproduced below.

\begin{lstlisting}
You classify whether a Hebrew tweet is conspiracy or non_conspiracy.

First, decide whether the tweet can be confidently classified from the text alone.
If yes, do not use any tools.

Only if the tweet is ambiguous (e.g., sarcasm, mockery, exaggeration, or unclear intent) and additional context could change the decision, you may use tools.

When using tools, choose them based on the missing information and use as few as necessary.

Stop using tools once you can confidently classify the tweet.
If a tool does not help, you may try a different tool.

Tool usage rules:
- tweet_info -> returns metadata (tweet date, text length, hashtags, URLs, posting source).
  Can be used to interpret hashtags and links, understand the timing of the tweet, and identify how it was posted (e.g., automated or unusual sources).
- user_info -> returns profile (bio, account age, followers, verification, language).
  Can be used when the author's identity can help understand intent. Can help identify ideological alignment, suspicious/bot-like users, or unusual account characteristics.
- previous_tweets -> returns earlier tweets by the same user with dates and text.
  Can be used to discover repeated narratives, consistent stance, or ongoing themes that indicate genuine belief rather than a one-off joke.
- ego_context -> returns top interacting users, their profile, and recent tweets.
  Can be used to understand the typical language and tone of the community (what is normal vs unusual). Neighbors' recent tweets can indicate what topics, claims, and narratives are currently discussed in the user's surrounding community, helping interpret an ambiguous tweet.

Return only valid JSON:
{
  "label": 0 or 1,
  "score": number between 0 and 1,
  "rationale": short explanation grounded in the tool outputs
}

Rules:
- score - conspiracy confidence (0 = not conspiracy, 1 = conspiracy)
- If you used tools, ground the rationale in those tool outputs and explicitly attribute evidence to tweet metadata, user profile, user history, or ego-context.
\end{lstlisting}

\subsection{Preloaded Context}

The preloaded conditions used the following system prompt:

\begin{lstlisting}
You classify whether a Hebrew tweet is conspiracy or non_conspiracy.

Classify using the tweet text and any additional context related to the tweet or user that is already provided in the user message.

Return only valid JSON:
{
  "label": 0 or 1,
  "score": number between 0 and 1,
  "rationale": short explanation grounded in the provided text and context
}

Rules:
- score - conspiracy confidence (0 = not conspiracy, 1 = conspiracy)
- If the provided context affects the decision, ground the rationale in that context and explicitly attribute the evidence to the relevant part of the provided context.
\end{lstlisting}

For preloaded conditions, the context was appended to the tweet text using the
following template before the common user-message template above was applied:

\begin{lstlisting}
<tweet_text>

Additional information that may help with classification:
<preloaded_context>
\end{lstlisting}

The available context was matched across the on-demand and preloaded
conditions:

\begin{table}[h!]
\centering
\small
\begin{tabular}{ll}
\hline
Condition & Available context \\
\hline
Tweet+User & \texttt{tweet\_info}, \texttt{user\_info} \\
History & \texttt{previous\_tweets} \\
Network & \texttt{ego\_context} \\
Full Context & all four sources \\
\hline
\end{tabular}
\caption{Context sources available in the matched on-demand and preloaded conditions.}
\label{tab:appendix_context_mapping}
\end{table}

\subsection{Two-Agent Debate}
\label{subapp:debate}

The debate condition used the same tweet input fields as the common input
template above and no external contextual tools.

\noindent\textbf{Analyst 1 system prompt.}

\begin{lstlisting}
You are Analyst 1 in a tweet-classification debate.
Follow the moderator's instruction exactly.

Task:
Classify tweets as:
- conspiracy
- non_conspiracy

For final decisions:
- Return valid JSON only
- label 1 = conspiracy
- label 0 = non_conspiracy
\end{lstlisting}

\noindent\textbf{Analyst 1 opening instruction.}

\begin{lstlisting}
Moderator: give your initial opinion.

Format:
- First line exactly: verdict: conspiracy
  or: verdict: non_conspiracy
- Then 1-3 short sentences explaining your reasoning.

Do not return JSON.
\end{lstlisting}

\noindent\textbf{Analyst 2 system prompt.}

\begin{lstlisting}
You are Analyst 2, a critical and skeptical reviewer in a tweet-classification debate.

Do not agree with Analyst 1 automatically, but do not disagree merely because the tweet is ambiguous. Evaluate the tweet independently and challenge Analyst 1 only when there is meaningful evidence that the classification may be incorrect.

Pay close attention to:
- sarcasm
- irony
- jokes or exaggeration
- mocking or non-literal tone

Only infer sarcasm or irony when there is clear evidence in the tweet. Do not treat extreme or unusual claims as sarcasm by default.

Distinguish between real conspiracy belief and language that imitates or mocks it.

Challenge the "conspiracy" label only when an alternative interpretation is clearly supported by the text.

Also challenge a non-conspiracy classification when the tweet appears to explain a significant social or political event through a secret plot by powerful actors.

Think independently and respond concisely in the required format.
Do not return JSON.
\end{lstlisting}

\noindent\textbf{Analyst 2 instruction.}

\begin{lstlisting}
Moderator: respond to Analyst 1.

Format:
- First line exactly: agreement: agree
  or: agreement: disagree
- Second line exactly: verdict: conspiracy
  or: verdict: non_conspiracy
- Then 1-3 short sentences explaining your reasoning.

Do not return JSON.
\end{lstlisting}

\noindent\textbf{Analyst 1 final instruction.}

\begin{lstlisting}
Moderator: read Analyst 2's response and make your final decision.

Decide whether to keep your original opinion or revise it based on Analyst 2's argument.
Only revise your decision if Analyst 2 provides a convincing reason.

Return EXACTLY one valid JSON object and nothing else:
{
  "label": 0 or 1,
  "score": number between 0 and 1,
  "rationale": "short explanation grounded in the discussion, reflecting whether Analyst 2 influenced your decision"
}

Rules:
- label 1 = conspiracy
- label 0 = non_conspiracy
- score = conspiracy confidence (0 = not conspiracy, 1 = conspiracy)
- The rationale should reflect your stance after considering Analyst 2
- No text before or after the JSON
\end{lstlisting}

\section{Original Hebrew Examples and Translations}
\label{app:hebrew_examples}

This appendix provides the original Hebrew texts, English translations, and posting dates for tweet examples quoted in the paper.

\subsection{Standalone Examples}

\noindent\textbf{Example H1: Pfizer storm tweet.}\par

\noindent\textbf{Date:} December 20, 2021.

\noindent\textbf{Original Hebrew:}
\begin{quote}
\centering
\includegraphics[width=\linewidth]{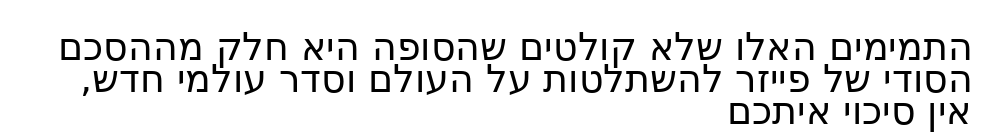}
\end{quote}

\noindent\textbf{English translation:}
\begin{quote}
``Those naive people do not realize that the storm is part of Pfizer's secret agreement to take over the world and create a New World Order. There is no hope for you.''
\end{quote}

\subsection{Discussion Examples}

\noindent\textbf{Example D1: Lemon and pharmaceutical concealment.}\par
\noindent\textbf{Date:} March 5, 2019.

\noindent\textbf{Original Hebrew:}

\begin{quote}
\centering
\includegraphics[width=\linewidth]{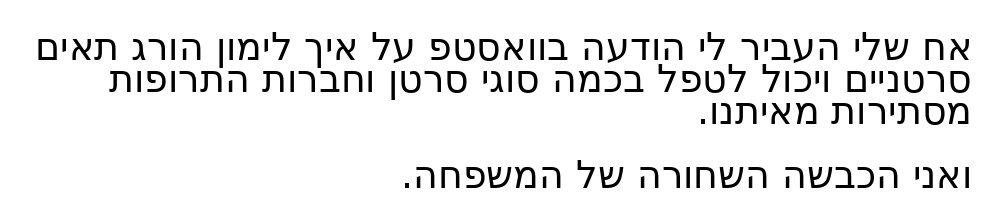}
\end{quote}

\noindent\textbf{English translation:}

\begin{quote}
``My brother sent me a WhatsApp message about how lemon kills cancer cells
and can treat several types of cancer, and how pharmaceutical companies are
hiding this from us. And I'm the black sheep of the family.''
\end{quote}

\noindent\textbf{Example D2: The Zombie Guide.}\par
\noindent\textbf{Date:} August 8, 2021.

\noindent\textbf{Original Hebrew:}

\begin{quote}
\centering
\includegraphics[width=\linewidth]{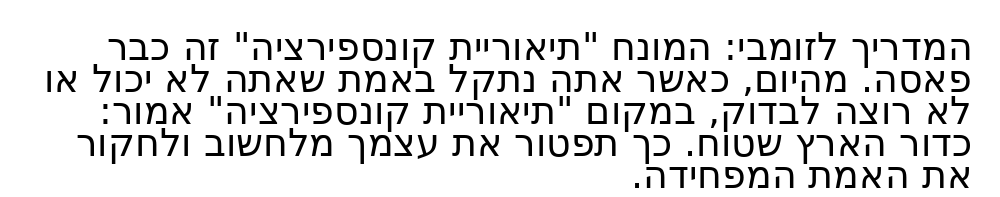}
\end{quote}

\noindent\textbf{English translation:}

\begin{quote}
``The Zombie Guide: The term `conspiracy theory' is already passé. From now
on, when you encounter a truth you cannot or do not want to investigate,
instead of saying `conspiracy theory,' say: the Earth is flat. That way, you
can avoid thinking about and investigating the frightening truth.''
\end{quote}

\subsection{Examples from Table~\ref{tab:motivating_examples}}

\noindent\textbf{Example T1: Chemtrails / crop-duster planes.}\par

\noindent\textbf{Date:} February 3, 2019.

\noindent\textbf{Original Hebrew:}
\begin{quote}
\centering
\includegraphics[width=\linewidth]{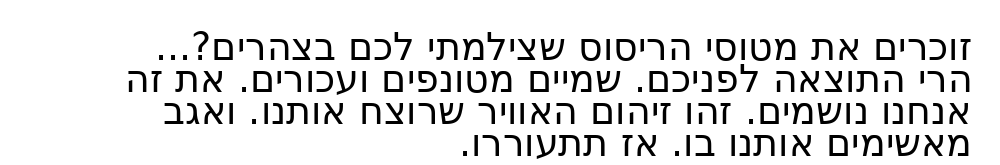}
\end{quote}

\noindent\textbf{English translation:}
\begin{quote}
``Remember the crop-duster planes I photographed this afternoon?... Here is the result. Dirty and murky skies. This is what we breathe. This is the air pollution that is killing us. And by the way, they blame us for it. So wake up.''
\end{quote}

\noindent\textbf{Example T2: Rabin assassination classroom report.}\par

\noindent\textbf{Date:} October 23, 2020.

\noindent\textbf{Original Hebrew:}
\begin{quote}
\centering
\includegraphics[width=\linewidth]{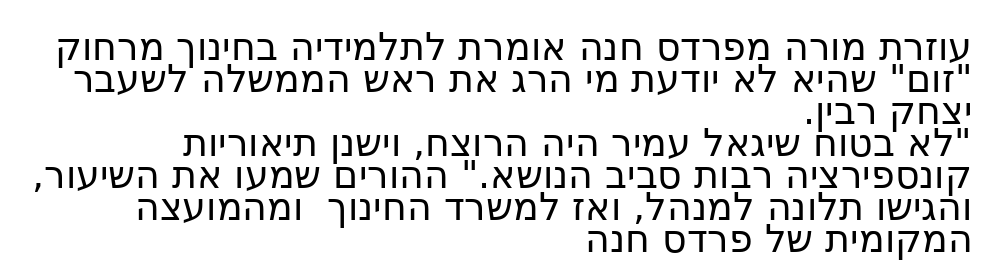}
\end{quote}

\noindent\textbf{English translation:}
\begin{quote}
``A teaching assistant from Pardes Hanna tells her students during remote learning on “Zoom” that she does not know who killed former Prime Minister Yitzhak Rabin.
“It is not certain that Yigal Amir was the murderer, and there are many conspiracy theories surrounding the issue.”
The parents heard the lesson and filed a complaint with the principal, then with the Ministry of Education and the Pardes Hanna local council.''
\end{quote}

\noindent\textbf{Example T3: COVID vaccine satire.}\par

\noindent\textbf{Date:} January 4, 2023.

\noindent\textbf{Original Hebrew:}
\begin{quote}
\centering
\includegraphics[width=\linewidth]{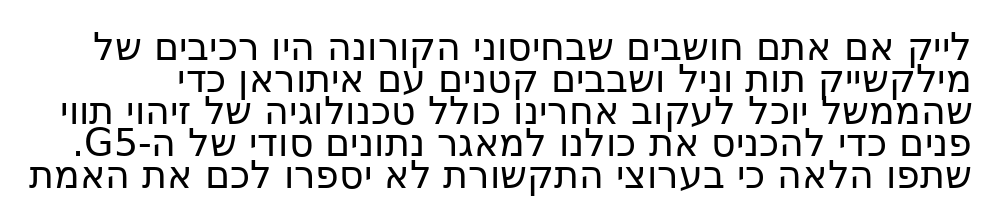}
\end{quote}

\noindent\textbf{English translation:}
\begin{quote}
``Like if you think the COVID vaccines contained strawberry-vanilla milkshake ingredients and tiny location-tracking chips, so the government could track us, including facial recognition technology, in order to put us all into a secret G5 database. Share this, because the media channels will not tell you the truth.''
\end{quote}

\noindent\textbf{Example T4: Hospital strike criticism.}\par

\noindent\textbf{Date:} August 31, 2021.

\noindent\textbf{Original Hebrew:}
\begin{quote}
\centering
\includegraphics[width=\linewidth]{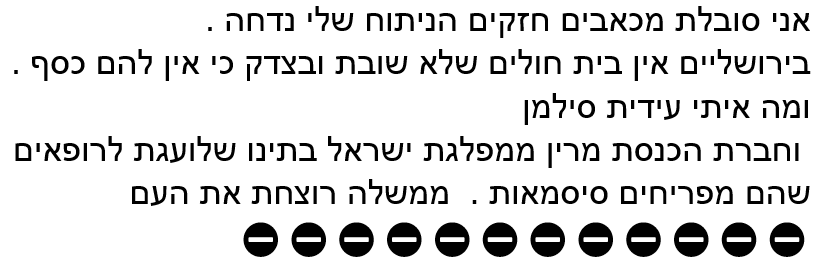}
\end{quote}

\noindent\textbf{English translation:}
\begin{quote}
``I am suffering from severe pain, and my surgery has been
postponed. In Jerusalem, every hospital is on strike, and rightly
so, because they have no money. And what about me, Idit Silman?
And Member of Knesset Marin from Yisrael Beiteinu, who mocks the
doctors by saying that they are merely making empty declarations.
A government that is murdering its people.''
\end{quote}

\noindent\textbf{Example T5: Climate policy lawsuit.}\par

\noindent\textbf{Date:} February 2, 2019.

\noindent\textbf{Original Hebrew:}
\begin{quote}
\centering
\includegraphics[width=\linewidth]{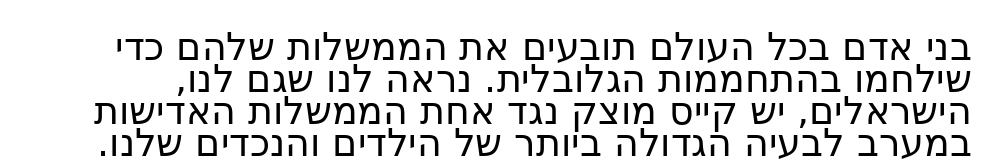}
\end{quote}

\noindent\textbf{English translation:}
\begin{quote}
``People around the world are suing their governments to force them to fight global warming. It seems to us that we Israelis also have a strong case against one of the most indifferent governments in the West regarding the greatest problem facing our children and grandchildren.''
\end{quote}

\section{Error Analysis Examples}
\label{app:error_examples}

\noindent\textbf{Example E1: Strong institutional criticism.}\par
\noindent\textbf{Date:} November 8, 2019.

\noindent\textbf{Original Hebrew:}

\begin{quote}
\centering
\includegraphics[width=\linewidth]{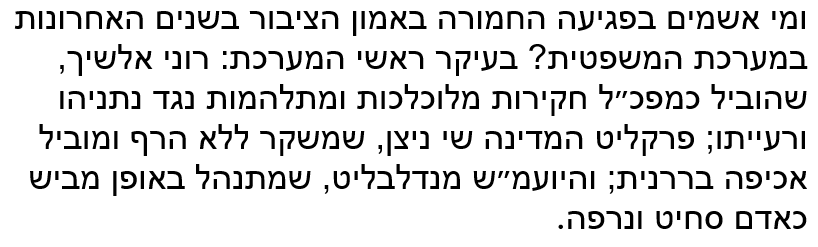}
\end{quote}

\noindent\textbf{English translation:}

\begin{quote}
``And who is responsible for the severe erosion of public trust in the
justice system in recent years? Mainly the system's leaders: Roni Alsheikh,
who as police commissioner led dirty and inflammatory investigations against
Netanyahu and his wife; State Attorney Shai Nitzan, who lies constantly and
pursues selective enforcement; and Attorney General Mandelblit, who conducts
himself disgracefully as a weak man susceptible to blackmail.''
\end{quote}

\noindent\textbf{Example E2: Satirical conspiracy language.}\par
\noindent\textbf{Date:} February 19, 2023.

\noindent\textbf{Original Hebrew:}

\begin{quote}
\centering
\includegraphics[width=\linewidth]{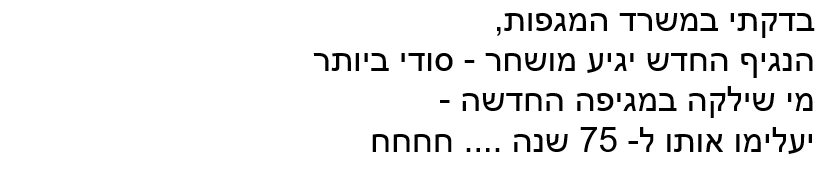}
\end{quote}

\noindent\textbf{English translation:}

\begin{quote}
``I checked with the Ministry of Pandemics.
The new virus will arrive redacted -- highly classified.

Whoever contracts the new epidemic will be made to disappear for
75 years ... hahaha.''
\end{quote}

\section{Conspiracy Topics and Retrieval Lexicon}
\label{app:key_terms}

This appendix lists the lexical resources used for rule-based candidate retrieval. The retrieval process uses general conspiracy cues and topic-specific rules. A matched tweet is not automatically labeled as conspiratorial.

\subsection{Conspiracy Topic Guide}

The candidate-retrieval process covers the following conspiracy topics and related narratives:

\begin{itemize}
    \item \textbf{Flat-earth claims:} The belief that Earth is a flat disc rather than a sphere, and that space agencies, governments, and scientists worldwide are colluding to fake evidence (like satellite images) of a round Earth.

    \item \textbf{Rabin assassination:} Beyond the established fact that Yigal Amir shot Israeli PM Yitzhak Rabin in 1995, various theories claim the Israeli security service (Shin Bet) orchestrated or deliberately allowed the assassination, sometimes alleging a broader plot involving multiple shooters or government complicity.

    \item \textbf{Yemenite children affair:} Refers to the disappearance of hundreds of infants, mostly from Yemenite Jewish immigrant families, in Israel during the 1950s. While official inquiries attributed most cases to poor record-keeping and high infant mortality, conspiracy theories allege a systematic, state-orchestrated program to kidnap the children and give them to Ashkenazi families or sell them abroad.

    \item \textbf{COVID conspiracy theories:} A range of claims including that the virus was deliberately engineered and released (sometimes tied to bio weapons labs), that the pandemic was exaggerated or fabricated to control populations, or that vaccines were designed to cause harm, implant tracking devices, or serve population-control agendas.

    \item \textbf{Medical and pharmaceutical control:} The belief that pharmaceutical companies, medical associations, and regulators suppress cheap or natural cures (for cancer, etc.) to protect profits from ongoing treatments, and/or that vaccines and medications are pushed for financial or population-control motives rather than genuine health benefits.

    \item \textbf{Global elites and the Illuminati:} The theory that a secretive group of powerful individuals or families (sometimes named after the historical 18th-century Bavarian Illuminati) secretly controls world governments, economies, and media to advance a hidden agenda, often toward a ``New World Order.''

    \item \textbf{QAnon:} A wide-ranging theory originating in 2017 alleging that a cabal of Satan-worshipping, child-trafficking elites in government, media, and business secretly runs the world, and that a hidden figure (``Q'') was leaking insider information about an anticipated reckoning against them.

    \item \textbf{Chemtrails:} The claim that visible trails left by aircraft are not ordinary condensation (contrails) but chemical or biological agents deliberately sprayed for purposes like weather control, population control, or geoengineering, hidden from the public.

    \item \textbf{9/11 conspiracy theories:} Various claims disputing the official account of the September 11, 2001 attacks, including that the U.S. government had foreknowledge and allowed them to happen (``LIHOP''), directly orchestrated them (``MIHOP''), or that the World Trade Center towers were brought down by controlled demolition rather than the plane impacts and fires.

    \item \textbf{Climate change denial/conspiracy:} Claims that anthropogenic climate change is a hoax or exaggeration fabricated by scientists, governments, or organizations for financial gain, political control, or to justify regulatory overreach, despite the strong scientific consensus supporting human-caused warming.

    \item \textbf{Moon-landing hoax:} The claim that NASA's Apollo moon landings (1969--1972) were staged, often in a film studio, to win the Cold War space race against the Soviet Union, and that supposed anomalies in photos and footage prove fabrication.

    \item \textbf{UFOs and aliens:} Theories ranging from claims that governments (especially the U.S.) have made contact with or recovered extraterrestrial spacecraft/beings and covered it up, to beliefs that aliens have influenced human history or secretly live among us.

    \item \textbf{The ``deep state'':} The belief that a hidden, unelected network of career bureaucrats, intelligence officials, and other entrenched interests secretly wields real power behind the scenes, undermining or controlling elected officials regardless of who holds office.
\end{itemize}
\subsection{Retrieval Rules}

\begin{table*}[t]
\raggedright
\textbf{General Conspiracy Cues}

General conspiracy cues are expressions that may indicate conspiratorial framing across multiple topics.

\centering
\includegraphics[width=\textwidth]{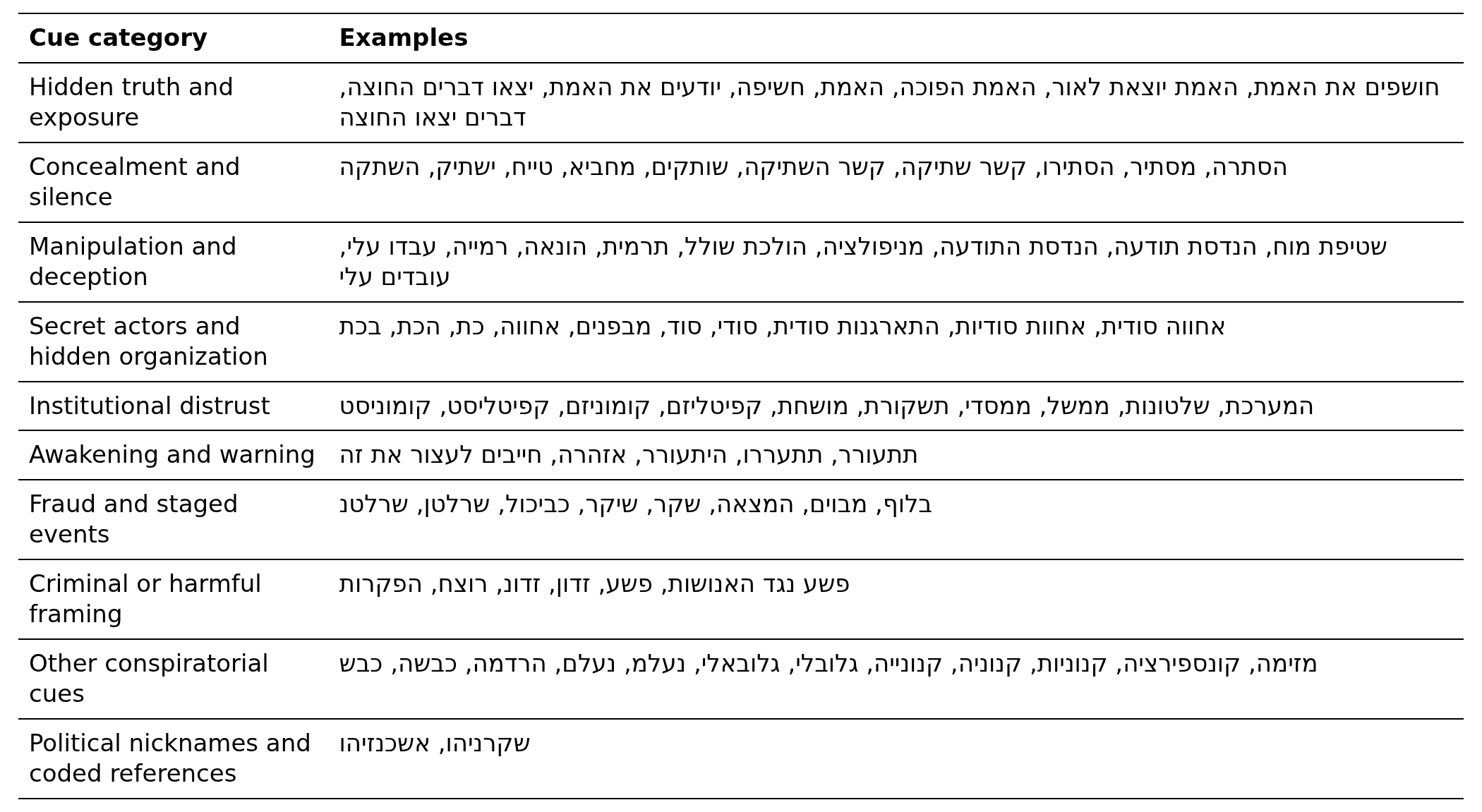}
\caption{Examples of general conspiracy cues used in candidate retrieval.}
\label{tab:general_conspiracy_cues}
\end{table*}

\begin{table*}[t]
\raggedright
\textbf{Topic-Specific Retrieval Rules}

Each topic contains three components: phrases, keywords, and threshold values. For each topic, a tweet is retrieved when it contains either enough topic-specific phrases or enough topic-specific keywords, and also contains enough general conspiracy cues.

\centering
\textbf{Flat Earth}

\includegraphics[width=\textwidth]{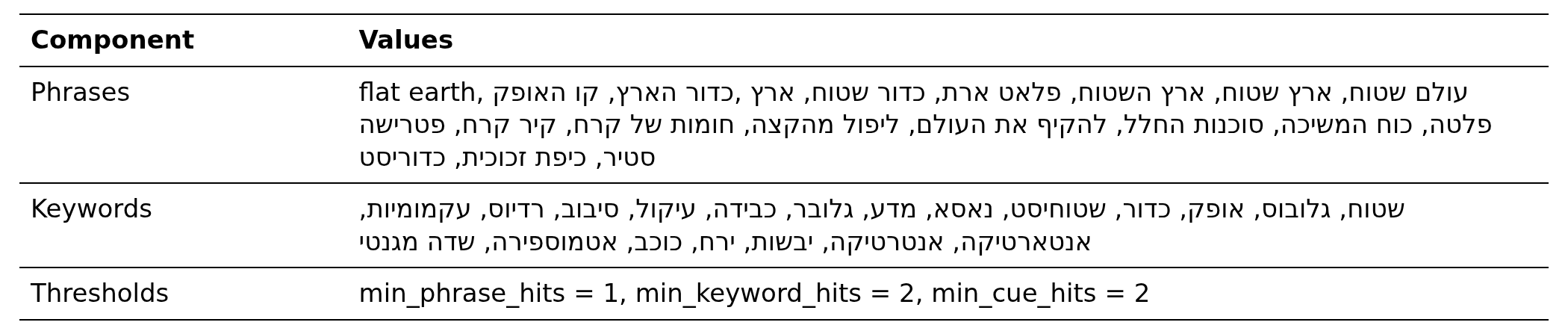}
\caption{Retrieval rule for the Flat Earth topic.}
\label{tab:rule_flat_earth}
\end{table*}

\begin{table*}[t]
\centering
\textbf{Yemenite Children Affair}

\includegraphics[width=\textwidth]{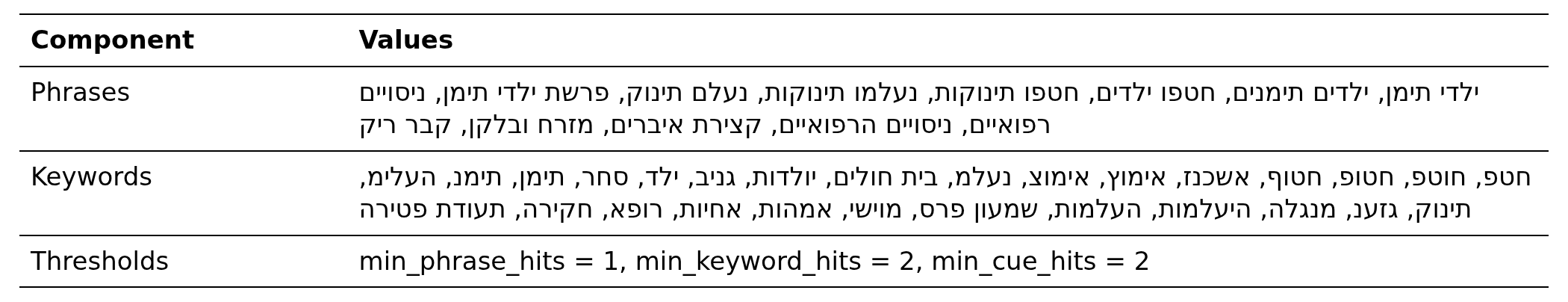}
\caption{Retrieval rule for the Yemenite Children topic.}
\label{tab:rule_yemenite_children}
\end{table*}

\begin{table*}[t]
\centering
\textbf{Rabin Assassination}

\includegraphics[width=\textwidth]{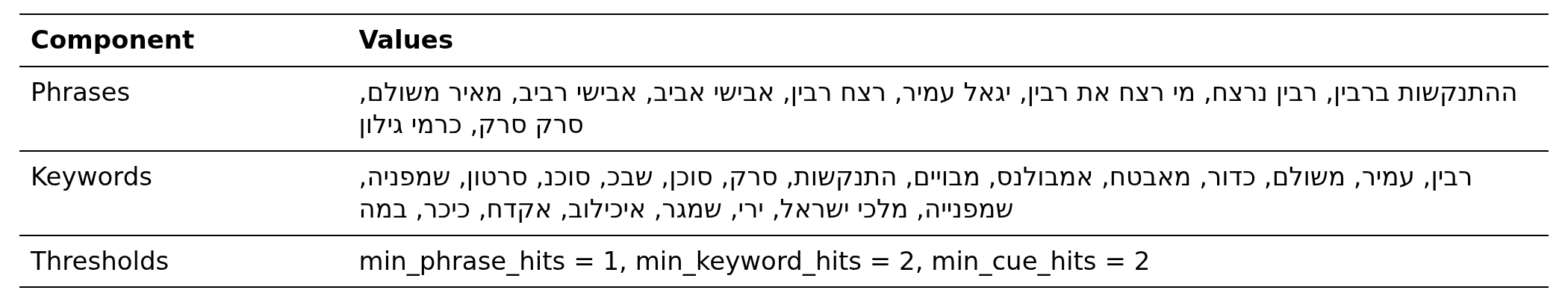}
\caption{Retrieval rule for the Rabin assassination topic.}
\label{tab:rule_rabin_assassination}
\end{table*}

\begin{table*}[t]
\centering
\textbf{Health and COVID Conspiracy}

\includegraphics[width=\textwidth]{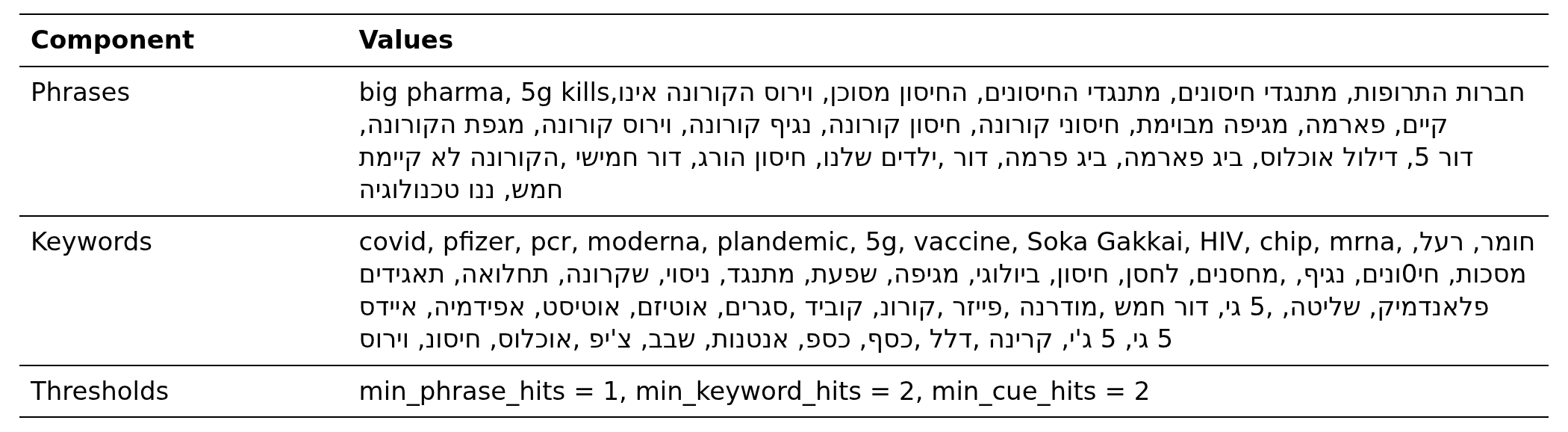}
\caption{Retrieval rule for the Health and COVID conspiracy topic.}
\label{tab:rule_health_covid}
\end{table*}

\begin{table*}[t]
\centering
\textbf{Illuminati and Global Elites}

\includegraphics[width=\textwidth]{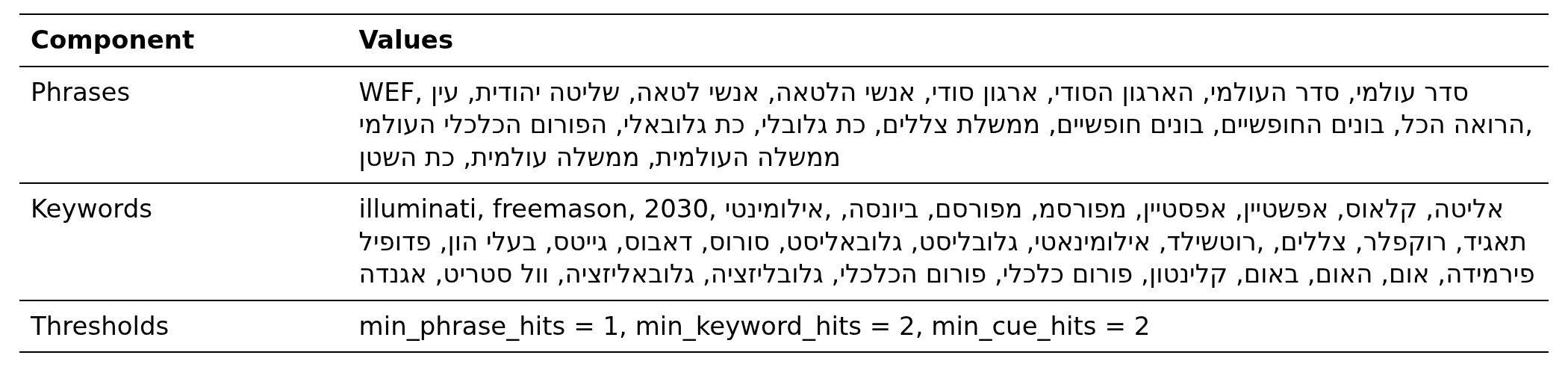}
\caption{Retrieval rule for the Illuminati and global elites topic.}
\label{tab:rule_illuminati}
\end{table*}

\begin{table*}[t]
\centering
\textbf{QAnon}

\includegraphics[width=\textwidth]{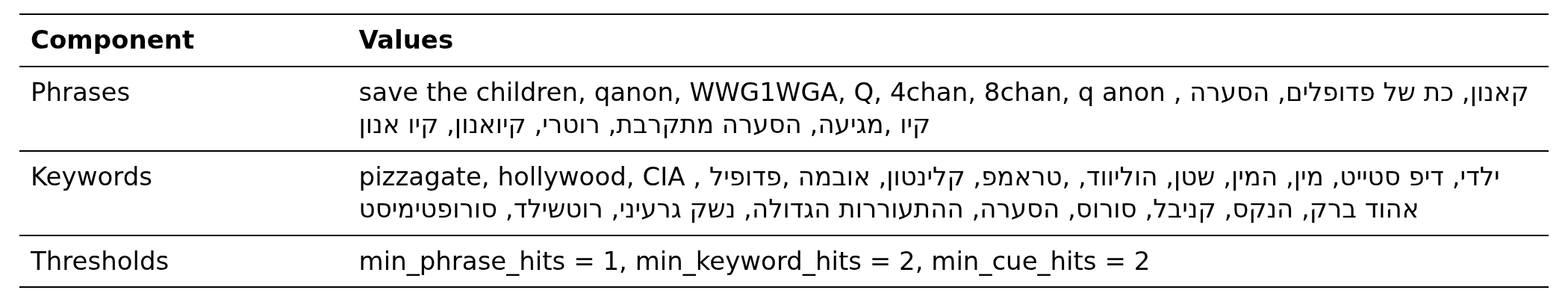}
\caption{Retrieval rule for the QAnon topic.}
\label{tab:rule_qanon}
\end{table*}

\begin{table*}[t]
\centering
\textbf{Chemtrails}

\includegraphics[width=\textwidth]{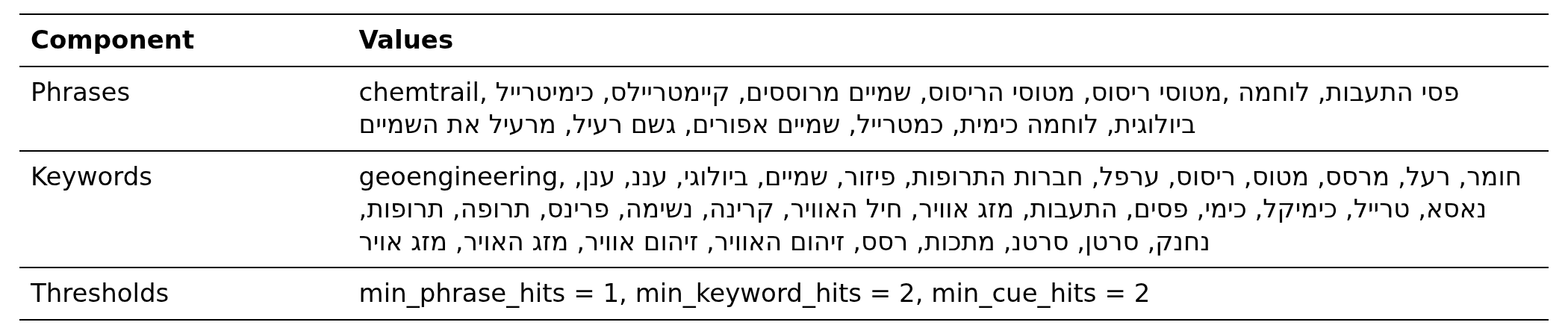}
\caption{Retrieval rule for the Chemtrails topic.}
\label{tab:rule_chemtrails}
\end{table*}

\begin{table*}[t]
\centering
\textbf{9/11}

\includegraphics[width=\textwidth]{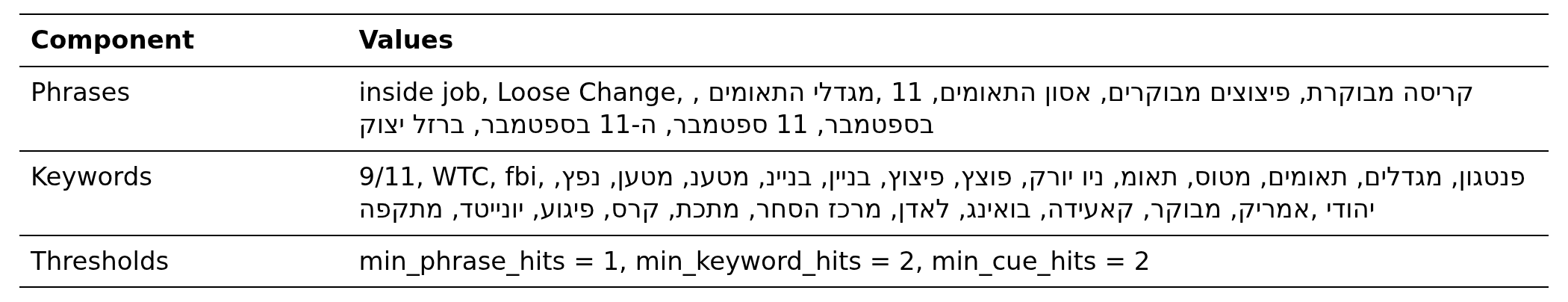}
\caption{Retrieval rule for the 9/11 topic.}
\label{tab:rule_nine_eleven}
\end{table*}

\begin{table*}[t]
\centering
\textbf{Climate Hoax}

\includegraphics[width=\textwidth]{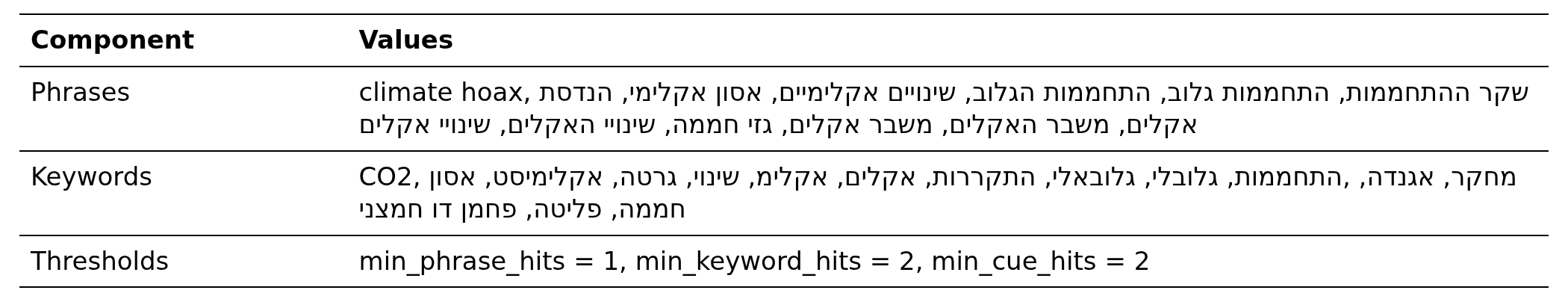}
\caption{Retrieval rule for the Climate Hoax topic.}
\label{tab:rule_climate_hoax}
\end{table*}

\begin{table*}[t]
\centering
\textbf{Moon Landing Hoax}

\includegraphics[width=\textwidth]{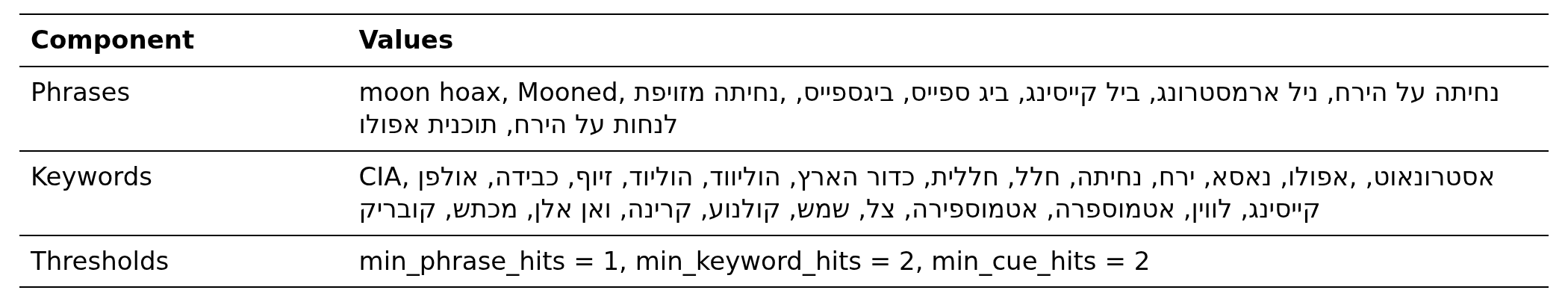}
\caption{Retrieval rule for the Moon Landing Hoax topic.}
\label{tab:rule_moon_hoax}
\end{table*}

\begin{table*}[t]
\centering
\textbf{UFOs and Aliens}

\includegraphics[width=\textwidth]{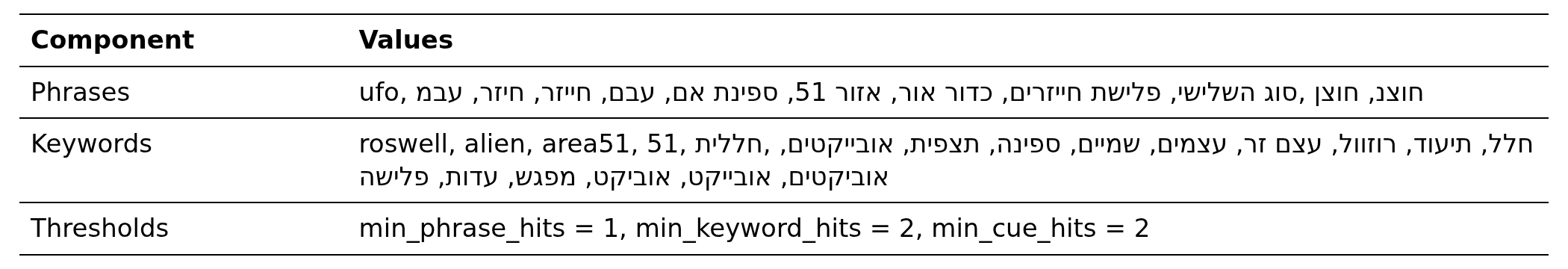}
\caption{Retrieval rule for the UFOs and aliens topic.}
\label{tab:rule_ufo_aliens}
\end{table*}

\begin{table*}[t]
\centering
\textbf{Deep State}

\includegraphics[width=\textwidth]{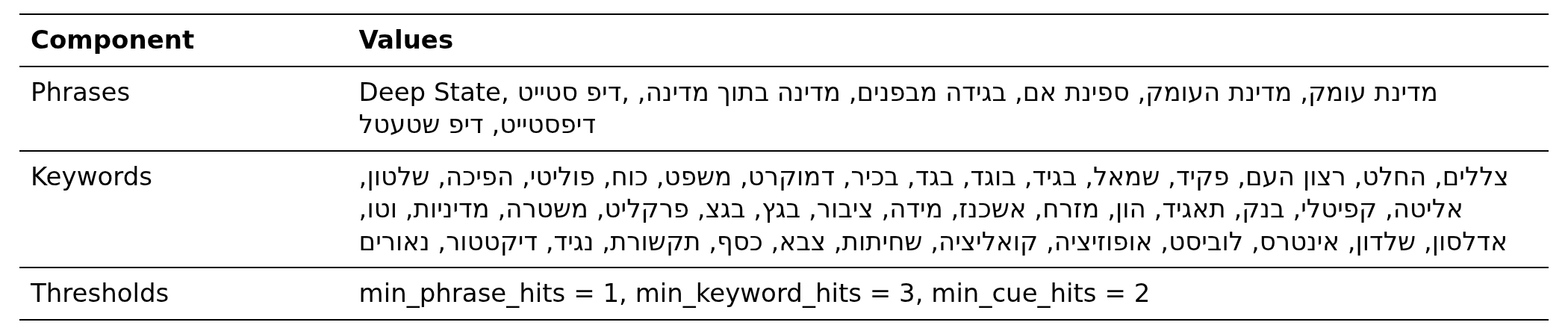}
\caption{Retrieval rule for the Deep State topic.}
\label{tab:rule_deep_state}
\end{table*}

\end{document}